\documentclass[10pt,twocolumn,letterpaper]{article}

\usepackage{cvpr}              %

\usepackage[dvipsnames]{xcolor}

\usepackage[export]{adjustbox}
\usepackage[accsupp]{axessibility}
\usepackage{graphicx}
\usepackage{amsmath}
\usepackage{amssymb}
\usepackage{booktabs}
\usepackage{stmaryrd}
\usepackage{multirow}
\usepackage{pifont}%
\newcommand{\cmark}{\ding{51}}%

\usepackage{colortbl} %
\usepackage{pgfplots} %
\usetikzlibrary{calc}
\pgfplotsset{compat=newest}

\newcommand{\acro}{PointBeV}

\definecolor{cvprblue}{rgb}{0.21,0.49,0.74}
\usepackage[pagebackref,breaklinks,colorlinks,citecolor=cvprblue]{hyperref}

\definecolor{lightgray}{gray}{0.85}

\definecolor{trainingcolor}{HTML}{e3f0fa}

\pgfplotsset{
    point plot style/.style={
        only marks,
        mark=*,
        nodes near coords,
        point meta=explicit symbolic,
    },
}

\usepackage[pagebackref,breaklinks,colorlinks]{hyperref}

\usepackage[capitalize]{cleveref}
\crefname{section}{Sec.}{Secs.}
\Crefname{section}{Section}{Sections}
\Crefname{table}{Table}{Tables}
\crefname{table}{Tab.}{Tabs.}

\def\appendixautorefname{App.}

\def\paperID{8488}
\def\confName{CVPR}
\def\confYear{2024}

\title{\acro{}: A Sparse Approach to BeV Predictions}

\author{
  \hspace{-0.4cm} Loick Chambon$^{1,2}$, \'Eloi Zablocki$^1$, Mickaël Chen$^1$, Florent Bartoccioni$^1$, Patrick Pérez\thanks{Work done at Valeo.ai.}$\hspace{0.15cm}^3$, Matthieu Cord$^{1,2}$ \\[0.2cm]
  $^1$ Valeo.ai, Paris, France \hspace{0.8cm} $^2$ Sorbonne Université, Paris, France \hspace{0.8cm} $^3$ Kyutai, Paris, France
}

\begin{document}

\maketitle

\begin{abstract}
   Bird's-eye View (BeV) representations have emerged as the de-facto shared space in driving applications, offering a unified space for sensor data fusion and supporting various downstream tasks.
However, conventional models use grids with fixed resolution and range and face computational inefficiencies due to the uniform allocation of resources across all cells.
To address this, we propose \acro{}, a novel sparse BeV segmentation model operating on sparse BeV cells instead of dense grids.
This approach offers precise control over memory usage, enabling the use of long temporal contexts and accommodating memory-constrained platforms.
\acro{} employs an efficient two-pass strategy for training, enabling focused computation on regions of interest.
At inference time, it can be used with various memory/performance trade-offs and flexibly adjusts to new specific use cases.
\acro{} achieves state-of-the-art results on the nuScenes dataset for vehicle, pedestrian, and lane segmentation, showcasing superior performance in static and temporal settings despite being trained solely with sparse signals.
We release our code with two new efficient modules used in the architecture: Sparse Feature Pulling, designed for the effective extraction of features from images to BeV, and Submanifold Attention, which enables efficient temporal modeling. The code is available at \url{https://github.com/valeoai/PointBeV}.
\end{abstract}

\setlength{\abovedisplayskip}{5pt}
\setlength{\belowdisplayskip}{5pt}

\section{Introduction}
\label{sec:intro}
Bird's-eye View (BeV) representations are now ubiquitously in driving applications.
Indeed, a top-view ego-centric grid is not only a convenient shared space for fusing inputs from multiple sensors \cite{li2022bevdepth,philion2020lift,ye2023fusionad,singh2023radar4bev}, but also a space-aware representation relevant for many downstream tasks such as detection \cite{li2022bevformer,liu2023bevfusion}, segmentation \cite{philion2020lift,fiery2021}, forecasting \cite{gu2023vip3d}, tracking \cite{yang2023bevtrack}, or planning \cite{dewangan2023uap-bev,hu2023uniad,qin2022unifusion}.
BeV segmentation encompasses a broad family of tasks such as 2D instance segmentation \cite{fiery2021}, 3D instance segmentation \cite{huang2023tri,zuo2023pointocc,jia2023occupancydetr}, occupancy forecasting \cite{stopnet,occupancy_flow_fields} and online mapping \cite{li2023bi-mapper}.
In this paper, we focus on {BeV segmentation from multiple cameras, in scenarios with or without past frames,}
respectively referred to as \emph{temporal} and \emph{static}.

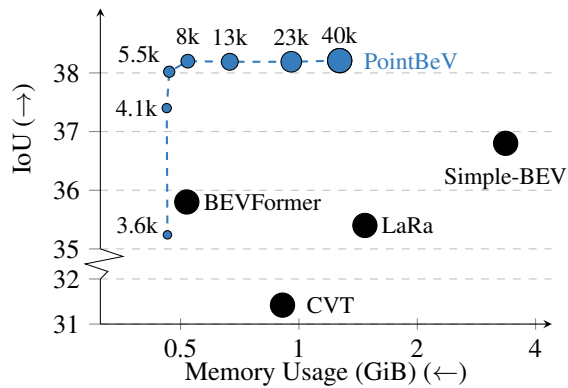
\begin{figure}[t]
\begin{tikzpicture}
\pgfmathsetmacro{\totalheight}{3cm}
\pgfmathsetmacro{\bottomheight}{\totalheight / 5}
\pgfmathsetmacro{\topheight}{4 * \bottomheight}

\pgfmathsetmacro{\gap}{4mm}

\pgfplotsset{
    point plot style/.style={
        only marks,
        mark=*,
        nodes near coords,
        point meta=explicit symbolic,
    },
}

\def\basePointSize{2pt} %

\newcommand{\scaledPoint}[6]{%
    \pgfmathparse{log10(#1)/log10(5)} %
    \edef\pointSize{\pgfmathresult*\basePointSize}
    \edef\temp{\noexpand\addplot[
        point plot style,
        mark options={fill=#4, scale=\pgfmathresult},
        every node near coord/.append style={#6},
    ] coordinates {
        (#2,#3) [#5]
    };}
    \temp
}

\begin{axis}[
    ymajorgrids, %
    grid style={dashed}, %
    tick align=outside,
    name=bottom axis,
    at={(0,0)},
    axis x line=bottom,
    axis y line*=left,
    every axis x line/.append style={-stealth}, %
    every axis y line/.append style={}, %
    ymin=31, ymax=32,
    ytick={31,32},
    yticklabels={31,32},
    xmin=320, xmax=4500,
    xmode=log, %
    log basis x={2}, %
    xtick={512,1024,2048,4096},
    xticklabels={0.5,1,2,4},
    xlabel style={yshift=5pt}, %
    scaled ticks=false, %
    height=\bottomheight,
    width=6cm,
    scale only axis,
    xlabel={Memory Usage (GiB) ($\leftarrow$)},
    nodes near coords align={vertical},
    every node near coord/.append style={
        font=\small,  %
    },
]

\scaledPoint{40}{930}{31.42}{black}{CVT}{anchor=west, xshift=2mm}

\end{axis}

\begin{axis}[
    name=top axis,
    ymajorgrids, %
    grid style={dashed}, %
    tick align=outside,
    at={($(bottom axis.north west)+(0,\gap)$)}, %
    anchor=south west,
    axis y line*=left,
    axis x line=none,
    ymin=35, ymax=39.1,
    xmin=320, xmax=4500,
    xmode=log, %
    log basis x={2}, %
    height=3.2cm,
    width=6cm,
    scale only axis,
    axis line style={-stealth},
    ytick={35,36,37,38},
    yticklabels={35,36,37,38},
    yticklabel pos=left,
    after end axis/.code={
        \draw [line width=0.1mm, black] 
        ($(axis cs:320,35)$) -- ++(3mm,-1mm) -- ++(-5mm,-1mm) -- ++(3mm,-1mm) -- ++(-1mm,-1mm);
    },
    ylabel={IoU ($\rightarrow$)},
    nodes near coords align={vertical},
    every node near coord/.append style={
        font=\small,  %
    }
]

\scaledPoint{40}{1300}{38.21}{cvprblue}{\textcolor{cvprblue}{PointBeV}}{anchor=west, xshift=2mm}
\scaledPoint{40}{1300}{38.21}{cvprblue}{40k}{anchor=south, yshift=1mm}
\scaledPoint{23}{979}{38.19}{cvprblue}{23k}{anchor=south,yshift=1mm}
\scaledPoint{13}{683}{38.19}{cvprblue}{13k}{anchor=south,yshift=1mm}
\scaledPoint{8}{534}{38.20}{cvprblue}{8k}{anchor=south,yshift=1mm}
\scaledPoint{5.5}{479}{38.02}{cvprblue}{5.5k}{anchor=south east}
\scaledPoint{4.1}{472}{37.40}{cvprblue}{4.1k}{anchor=east}
\scaledPoint{3.6}{474}{35.24}{cvprblue}{3.6k}{anchor=south east, yshift=-1mm}

\addplot[
    color=cvprblue,
    dashed,
    thick
] 
coordinates {
    (1300,38.21)
    (979,38.19)
    (683,38.19)
    (534,38.20)
    (479,38.02)
    (472,37.40)
    (474,35.24)
};

\scaledPoint{40}{3437}{36.8}{black}{Simple-BEV}{anchor=north, yshift=-2mm}
\scaledPoint{40}{531}{35.8}{black}{BEVFormer}{anchor=west, xshift=1mm}
\scaledPoint{40}{1507}{35.4}{black}{LaRa}{anchor= west, xshift=1mm}

\end{axis}
\end{tikzpicture}

\centering

\vspace{-0.15cm}
\caption{\textbf{BeV vehicle IoU vs. memory footprint on nuScenes \cite{caesar2020nuscenes} validation set.} Models are evaluated without visibility filtering (i.e all annotated vehicles are considered) at resolution $224 \times 480$. The memory consumption is calculated using a 40GB A100 GPU. The size of a dot represents the number of BeV points being evaluated, the smaller the better.
\acro{} has the capacity to explore various trade-offs between efficiency and performance by varying the number of points being {considered}. %
The remaining points are considered as zeros in the final prediction.
Using \acro{} we can achieve state-of-the-art performance with only a small portion of the points and without %
losing performance.
}

\label{fig:iou_vs_mem}
\end{figure}

BeV representations are usually implemented using grids of fixed resolution and range \cite{fiery2021,li2022bevformer,harley2022simple,li2023powerbev,fang2023tbpformer}.
This limits their efficiency in terms of compute, even more clearly when considering temporal tasks, where aggregating past frames for long horizons can be a very costly endeavor. 
Departing from these dense BeV grid approaches, we present in this paper \acro{}, a camera-based BeV segmentation model that operates on sparse BeV features.  Our approach offers control over the model's memory usage by restricting the number of points considered, and enables adaptive focus on specific regions of interest. Notably, we develop two modules for efficient sparse operations: the \emph{Sparse Feature Pulling} module, which retrieves features from multiple cameras using sparse coordinates, and the \emph{Submanifold Attention} module, adapted from submanifold attention \cite{lai2022stratified}, to handle our temporal aggregation strategy.
At test time, \acro{} operates in a low-compute regime without retraining and possesses the flexibility to adjust computation based on varying use cases.

Despite being trained solely with sparse signals, \acro{} attains state-of-the-art results in classic BeV segmentation tasks --- vehicle, pedestrian, and lane segmentation --- on the nuScenes dataset \cite{caesar2020nuscenes}, both in static and temporal settings. The performance and flexibility of \acro{} are demonstrated in \autoref{fig:iou_vs_mem}.

\noindent Our main contributions are as follows:
\begin{itemize}
    \itemsep0em
    \item We introduce a sparse BeV paradigm for BeV segmentation task, allowing to operate in memory-constrained settings without performance drop.
    \item We introduce two sparse modules: a \textit{Sparse Feature Pulling} module
    {for efficient features extraction from images to BeV and, in temporal settings, \textit{Submanifold Attention} for efficient temporal aggregation allowing to process an extended number of past frames.}
    \item We develop specific training, inference for this sparse BeV paradigm. The sparse aspect of \acro{} broadens its utility. Without retraining, it can balance at test-time efficiency and precision, or exploit at additional information like LiDAR point clouds and HD maps.
    \item Our \acro{} reaches state-of-the-art results for
    vehicle, pedestrian and lane segmentation %
    under different visibility settings and at different image resolutions.
\end{itemize}

\section{Related Work}
\label{sec:rw}

\begin{figure*}
    \centering
    \includegraphics[width=\textwidth]{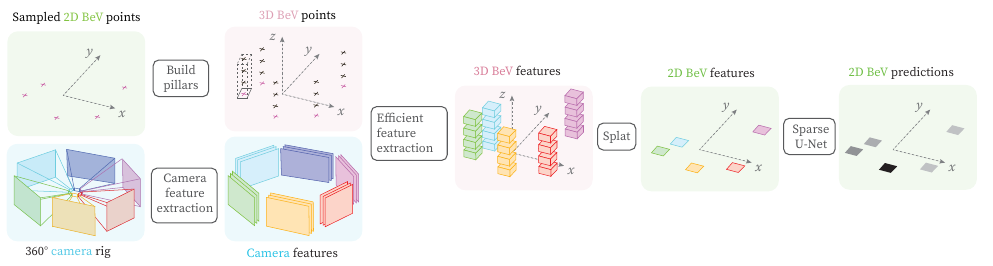}
    \vspace{-0.55cm}
    \caption{\textbf{\acro{} architecture.}
    As a sparse method, \acro{} is trained using local predictions, only for sampled 2D points provided as inputs. The selection of those points during training and at test time is illustrated in \autoref{fig:coarse_fine}.
     The points of interest are lifted to form 3D pillars, with each 3D point pulling visual features. To achieve this, \acro{} incorporates an efficient feature extraction process through a \textit{Sparse Feature Pulling} module, illustrated in the `efficient feature extraction' block and further explained in \autoref{sec:model:sparse_feature} and \autoref{fig:feature_extraction}.
     The obtained 3D BeV features are then flattened onto the 2D BeV plane and processed using a sparse U-Net with task-dependent final heads, generating local BeV predictions.
     {For training, we only need sparse signals.}
     {At} %
     test time, points that have not been sampled are 
     {set to zero.} %
    }
    \label{fig:archi}
    \vspace{-0.3cm}
\end{figure*}

\paragraph{Vision-based BeV Segmentation.}

BeV offers a convenient space for aggregating multiple cameras \cite{philion2020lift} and as such has become one of the main representations for autonomous driving \cite{chitta2023transfuser,chitta2021neat,hu2023uniad,xu2023endtoendforecasting,occupancy_flow_fields,sarlin2023snap}. 
Nevertheless, changing perspective from camera images to build BeV maps is challenging, involving depth estimation and 3D geometry \cite{li2022bevdepth}.

One approach learns the projection entirely from data, typically using a learnable-query-based cross-attention mechanism.
Typically, geometric embeddings such as frustum 3D coordinates \cite{liu2022petr,liu2022petrv2,Xiong2023CAPE} or ray vectors \cite{zhou2022crossview,bartoccioni2022lara,Pan_2023_CVPR}, are injected to enrich image features with 3D knowledge.  

Nonetheless, these methods lack interpretability as the mapping from perspective view to BeV is only implicit. 
Instead, some methods include a differentiable geometric projection in their approach.
In the seminal Lift-Splat-Shoot (LSS)~\cite{philion2020lift} work, a depth probability is estimated for each feature of the image, allowing to `lift' each feature to form a 3D voxel grid.
Then, the per-camera voxels are simply flattened (`splat') onto a common BeV grid.
However, for this pipeline to work, precise depth estimation is essential to accurately project image features to their corresponding 3D location \cite{simonelli2021missingconfidence}, 
and subsequent works focused on improving this particular aspect \cite{zhou2022matrixvt,li2022bevdepth,li2022bevstereo,li2023bevstereo,yang2023parametric,xie2022m2bev}. 
For instance, BEVDepth~\cite{li2022bevdepth} introduces LiDAR supervision while 
BEVStereo~\cite{li2022bevstereo,li2023bevstereo} leverages stereo depth-estimation techniques, both aiming at helping the depth estimation.
Parametric-Depth~\cite{yang2023parametric} instead imposes a Laplacian distribution as a prior to reduce the ambiguity of the estimation.%

Unlike aforementioned methods, BEVFormer~\cite{li2022bevformer} and Simple-BEV~\cite{harley2022simple} `pull' features to the BeV space without relying on estimated depth maps.
Instead, they define a 3D volume of coordinates over the BeV plane, project each 3D coordinate into the camera images, and pull the image features neighboring the resulting locations.
The image features are thus not precisely located in the BeV but instead spread over possible locations.
This greatly simplifies the process, making it both more efficient and robust to projection errors.
In our work, we build upon these `feature pulling' methods
and we further improve their efficiency and flexibility using sparse representations.

\vspace{.1cm}\noindent\textbf{Temporal Modelisation.}
A variety of temporal fusion methodologies has been explored, particularly warp-based techniques \cite{huang2022bevdet4d,li2022bevformer,fiery2021,hu2023fusionformer,akan2022stretchbev,zhang2022beverse,li2023powerbev,han2023videobev,park2023solofusion}.
These methods typically factor in ego-motion to spatially align historical BeV grids to the current BeV.
This facilitates the aggregation of multiple temporal frames by simply stacking the aligned BeV or with relevant neural modules such as deformable attention \cite{li2022bevformer} or Conv-RNNs \cite{zhang2022beverse,akan2022stretchbev}.
In addition to the warping relative to the ego-motion, BEVerse \cite{zhang2022beverse} and PowerBEV \cite{li2023powerbev} also explicitly model the other vehicles' instance flow and displace them to their predicted current location.
However, naive BeV map warping techniques only retain the spatial region from past data that aligns with the current BeV, discarding potentially valuable information from other regions. 
This hinders long-term temporal fusion, as the overlap in BeV decreases with the ego-car's movement \cite{qin2022unifusion,fang2023tbpformer}.
To circumvent these issues, methods like UniFusion \cite{qin2022unifusion} and TBP-Former \cite{fang2023tbpformer} propose to project the current BeV to the previous camera frames, instead of warping past BeV to the present one.
Still, the aggregation now suffers from the opposite problem, as many past frames, especially far away in time, would only contain little information relevant to the current location.
Yet, they would still be fully computed and propagated through the network.

Our temporal sparse approach allows us to control the compute by removing the less relevant points before the temporal aggregation, and thus to attain long past horizons.
Note that some of the works cited in this section, do not tackle BeV segmentation tasks \cite{fang2023tbpformer}.

\section{\acro{}}
\label{sec:model}
Our method \acro{} focuses on efficiently converting perspective camera views to BeV. The approach is outlined in three key components.
First, in \autoref{sec:model:sparse_feature}, we detail the efficient transformation of perspective camera views to BeV using sparse techniques, introducing our custom \textit{Sparse Feature Pulling} module.
Second, we introduce a two-stage `coarse/fine' learning strategy in \autoref{sec:model:coarse_fine} that explores and focuses on regions of interest, enabling the training of the model with significantly fewer points.
Third, in \autoref{sec:model:temporal}, we leverage the sparse nature of \acro{} to develop a submanifold temporal attention aggregation strategy that focuses on regions of interest and mitigates ambiguities associated with low visibilities.
Lastly, in \autoref{sec:model:inference}, we present sparse sampling strategies that enable the exploration of different efficiency/accuracy trade-offs at inference time.
An overview of the {architecture is shown} in \autoref{fig:archi}.

\begin{figure}[t]
    \centering
    \includegraphics[width=\columnwidth]{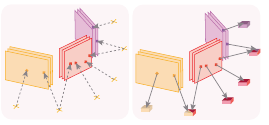}
    \vspace{-0.60cm}
    \caption{\textbf{Sparse Feature Pulling and Camera Fusion.} 3D BeV points are projected into the localized camera features (left). From there, camera features are bilinearly interpolated to obtain the 3D BeV features at this position (right).
    Where previous methods project points onto all the cameras regardless of their visibility, or pad the number of points so that there are as many per camera,
    {we conduct feature pulling, for each camera, only on the visible 3D points.}
    If a point is visible to a single camera, the feature pulling is done only within the corresponding feature volume. %
    }
    \label{fig:feature_extraction}
\end{figure}

\begin{figure*}
    \centering
    \includegraphics[width=\textwidth]{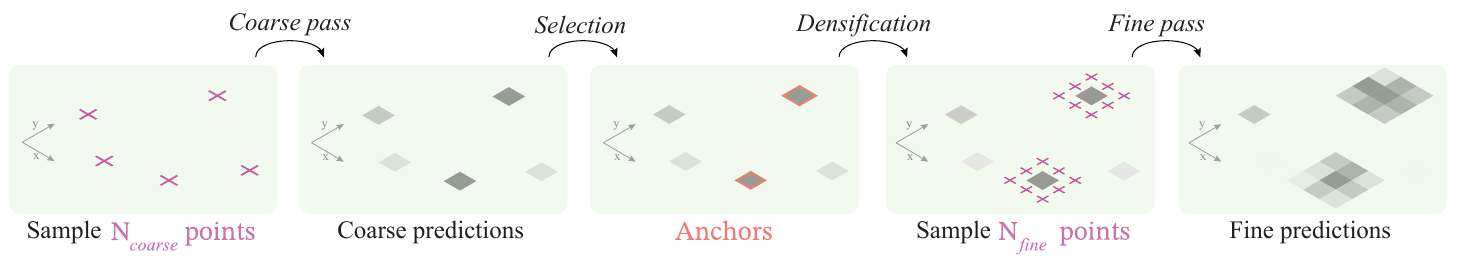}
    \vspace{-0.3cm}

    \newlength{\squareSize}
    \setlength{\squareSize}{0.15\textwidth} %

    \newlength{\spacing}
    \setlength{\spacing}{0.01cm} %

    \begin{tabular}{@{} ccc @{}}
        \includegraphics[width=\squareSize]{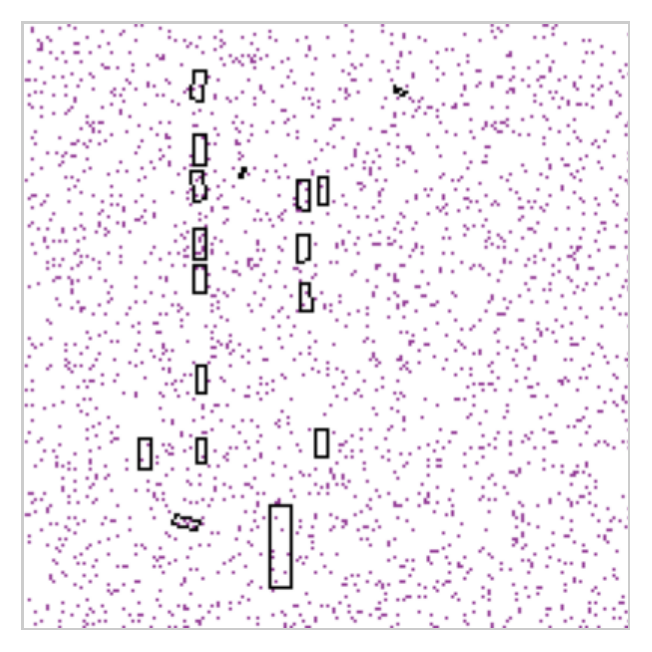}
        \hspace{\spacing}
        \includegraphics[width=\squareSize]{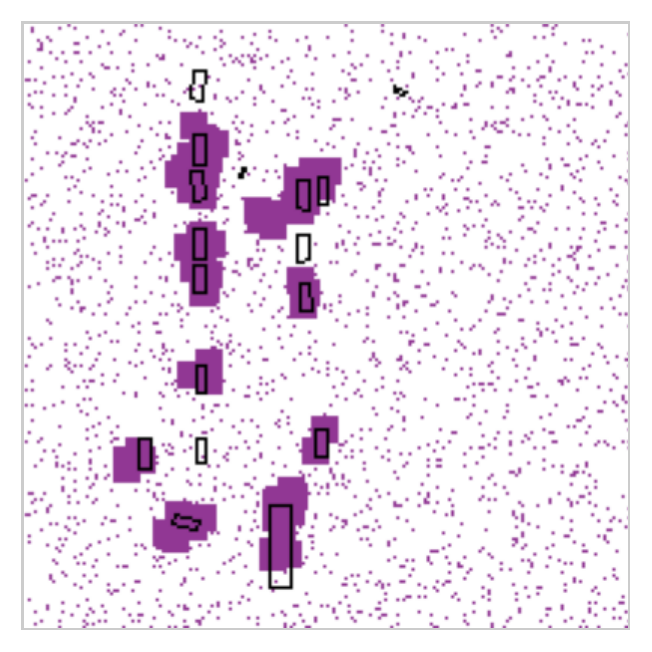} &
        \includegraphics[width=\squareSize]{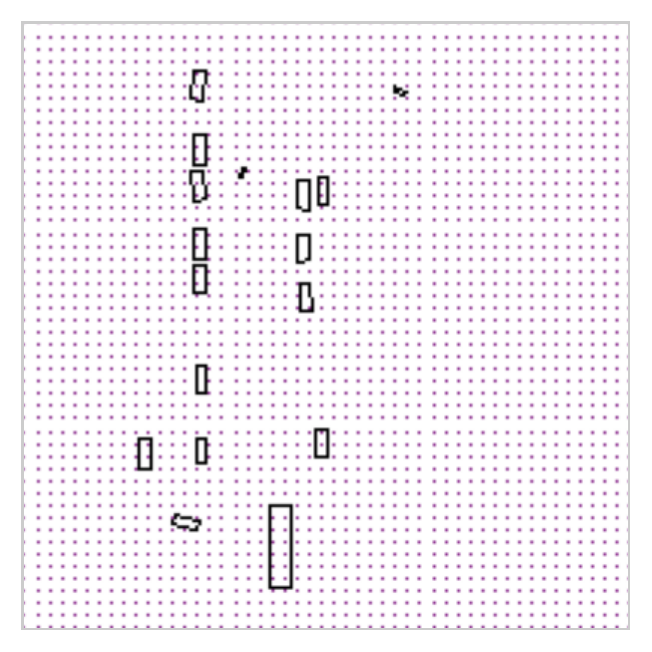}
        \hspace{\spacing}
        \includegraphics[width=\squareSize]{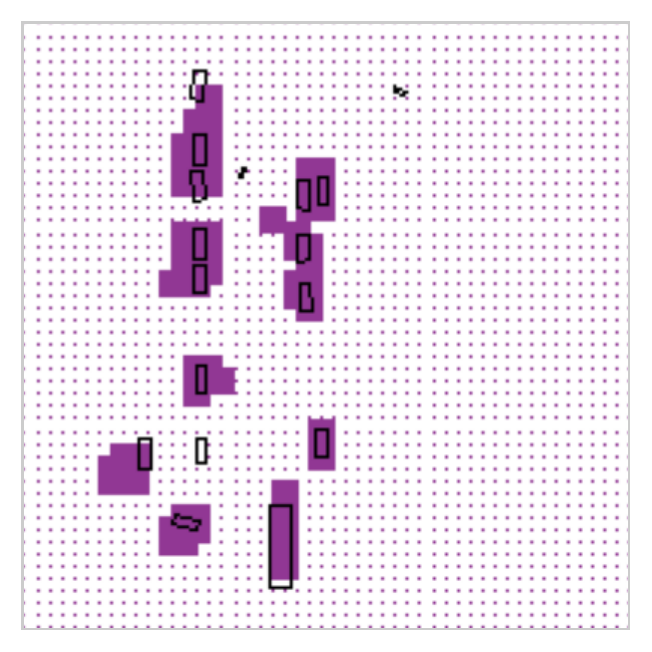} &
        \includegraphics[width=\squareSize]{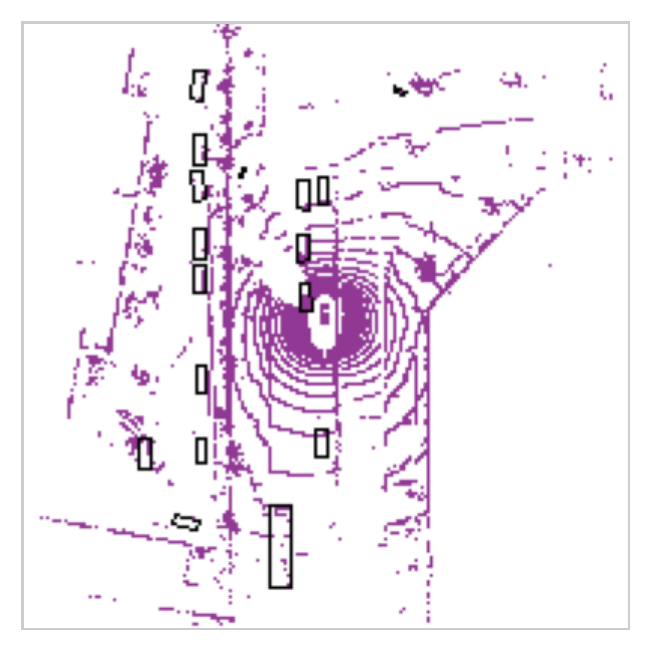}
        \hspace{\spacing}
        \includegraphics[width=\squareSize]{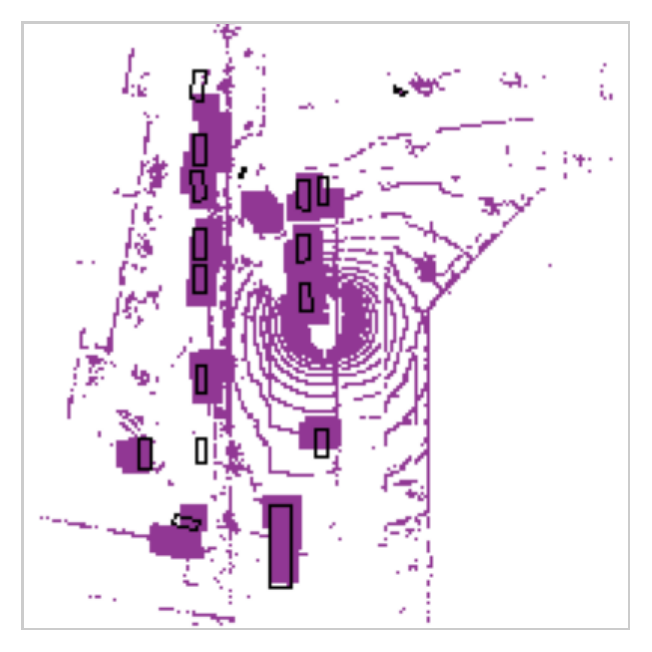} \\
         Random coarse and densified (training) &  Regular coarse and densified & LiDAR coarse and densified \\
    \end{tabular}

    \vspace{0.15cm}
    \caption{\textbf{Illustration of the `coarse' and `fine' passes.}
    Top row: given sampled BeV points, predictions are made at these locations in the `coarse pass'.
    {We select highest logits points as `anchors'.}
    Around these anchors, points are densely sampled using a kernel of size $k_\textit{fine} \times k_\textit{fine}$ {($3\times3$ in this vizualisation)}. %
    Then the `fine pass' %
    {provides}
    predictions for these points.
    {The networks (\autoref{fig:archi}) are shared between passes, and the camera feature extraction is only done once as the features don't change.}
    This figure illustrates both the training and the inference stages, and we stress non-visible differences between these two.
    \emph{During training}, (1) the coarse points are typically randomly sampled from a uniform distribution, and (2), the top $N_\textit{anchor}$ activations are selected as anchors. 
    \emph{During inference}, (1) the coarse points are sampled using different strategies such as the subsampled pattern (see \autoref{sec:model:inference}), and (2) points having a score above the threshold $\tau$ are selected as anchors.
    To evaluate the entire dense BeV, {we instead make a single pass with all BeV points.} %
    The bottom row displays \textcolor{violet}{sampling masks} for three different sampling strategies, with the ground-truth vehicles' bounding boxes delineated in black %
    for visualization.
    }
    \label{fig:coarse_fine}
\end{figure*}
\subsection{Sparse Feature Propagation}
\label{sec:model:sparse_feature}

\acro{} is a sparse approach to BeV segmentation.
Specifically, given a set of 2D BeV locations, also called `points', \acro{} predicts the occupancy state of these points from the camera images.
This contrasts with traditional grid-based approach that predicts the entire BeV.
The next subsections detail the selection of these 2D BeV locations during both training and inference, whereas here, we present the network architecture working with points.

Formally, for a given camera $C_i$ out of the $N_\textit{cam}$ cameras, we extract visual features using any backbone network.
These features are noted ${\mathcal{I}_i \in \mathbb{R}^{C\times H \times W}}$, where ${C, H, W \in \mathbb{N}}$ represent the number of channels, height and width of the feature volumes, respectively.
For each 2D BeV point $(x, y)$, we classically construct a pillar composed of 3D points evenly spaced vertically in the BeV space $p = \{p_z = (x, y, z)\}_{z \in Z}$ where $Z$ is the evenly discretized vertical axis \cite{lang2019pointpillars}.
Then, for any given 3D pillar point $p_z$, we define the set of camera indices $C(p_z) := \{i \mid C_i \sphericalangle  p_z\}$ representing the cameras that have the pillar point $p_z$ in their field of view (`is seeing' is noted `$\sphericalangle$').
Typically, in most multi-camera datasets, a point is visible to one or two cameras depending on the overlapping areas.
The 3D pillar point $p_z$ is then projected into the camera feature volumes $\{\mathcal{I}_i\}_{i \in C(p_z)}$, and a bilinear interpolation is performed to calculate the features corresponding to the 3D point.
We refer to the joint operations of projecting the 3D pillar points to the cameras, and extracting and interpolating the corresponding image features as \emph{Sparse Feature Pulling}.
When the pillar point falls in several cameras, the feature pulling is conducted separately in each camera feature volume and the features are then averaged.
This part of the feature propagation pipeline is illustrated in \autoref{fig:feature_extraction}.

We should note that contrary to most recent BeV approaches \cite{harley2022simple,li2022bevformer}, we do not project our points on cameras outside of their visibility field during feature pulling.
As in most of the autonomous driving multiple-camera datasets containing six cameras, each of them seeing about $\sim$16\% of the BeV space, we avoid $\sim$84\% of dispensable calculations that are often still performed in approaches.
To achieve this feat, we implement a \emph{Sparse Feature Pulling} module that can be used to replace any multi-view feature pulling scenario. We detail it and report its efficiency in \autoref{sec:supp:sparse_feature_pulling}.

Once features are obtained for considered pillar points, they are flattened onto the 2D BeV plane and processed using a sparse U-Net, generating local BeV segmentation predictions.
The sparse U-Net is a sparse adaptation of the classic U-Net model used in previous approaches \cite{harley2022simple,bartoccioni2022lara} containing commonly used ResNet blocks \cite{he2016resnet}.

\subsection{Coarse and fine training}
\label{sec:model:coarse_fine}

Given the dense nature of segmentation tasks, previous approaches treat the BeV grid as an image.
{This approach necessitates a large amount of memory, typically preventing training models on small GPUs, and using large temporal windows, fine-grain resolution, or long BeV ranges.}
Instead, during learning, we opt to subsample points from the BeV grid in the training phase.
However, naively tackling the dense segmentation task with these points results in training instabilities. %
To address this, we take inspiration from sampling strategies used in the NeRF literature \cite{mildenhallNeRF} and conceive a two-stage `coarse/fine' learning approach.
The goal is threefold: enhancing performance by focusing on discriminative regions, stabilizing the training process, and controlling memory usage during training.

This `coarse/fine' training strategy, as illustrated in \autoref{fig:coarse_fine}, involves two distinct passes.
The \emph{coarse pass} aims to cover the space as efficiently as possible, while the subsequent \emph{fine pass} focuses on areas identified as relevant by the coarse pass.
In the coarse pass, we use a uniform sampling strategy to draw a fixed number $N_\textit{coarse}$ of points that we forward in \acro{} (\autoref{fig:archi}).
Then, we select anchor points as the $N_\textit{anchor}$ points with the highest logits.
For the fine pass, we consider these anchor points and their neighbors in a square window of size $k_\textit{fine} \in \mathbb{N}$, a strategy we term \textit{densification}.
Among all the obtained densified points, we then keep $N_\textit{fine}$ points that we forward through the network for predictions (\autoref{fig:archi} again).
The outputs from both passes are merged before computing the cross-entropy, considering only the selected points.
This strategy allows us to oversample regions of interest while reducing the total amounts $N_\textit{point} := N_\textit{coarse} + N_\textit{fine}$ of points that are processed.%
Typically, we split equally the total number of points between the two stages: ${N_\textit{coarse} = N_\textit{fine}}$ to maintain a balance between exploration and refinement. %

\subsection{Sparse temporal model}
\label{sec:model:temporal}

Leveraging past frames to build BeV maps can help to discern static from dynamic objects, understand traffic patterns, and alleviate ambiguities 
{caused by} transient obstructions, visual artifacts, or depth uncertainty. 
For temporal aggregation, we take inspiration from the strategy used in BeV detection \cite{qin2022unifusion} to project the current locations of interest into past images, {and} we adapt it to a sparse setting.

Let us consider a sequence of $T \in \mathbb{N}$ frames encompassing both past and present contexts, with the goal of predicting the BeV map of the present moment.
To achieve this, the coordinates of sampled points in the current map are projected into the cameras of past moments.
This process yields the features of points in the current map for each time frame.
However, instead of aggregating all points, we selectively retain past points based on their logit values, introducing a temporal threshold $\tau_\textit{temp} \in [0,1]$.
Points exceeding this threshold are preserved, while those below it are discarded. 
{This results in a sparse BeV per timestep.}

\begin{figure}
    \centering
    \includegraphics[width=0.8\columnwidth]{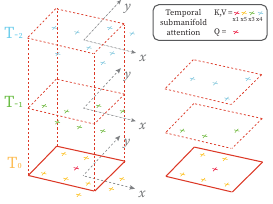}
    \caption{\textbf{Illustration of the `Submanifold Temporal Attention' module.} Our module performs an attention between a query point (colored in \textcolor{red}{red}), at the center of a spatio-temporal neighborhood (\textcolor{red}{red} dotted lines and complete parallelepiped). The points inside this neighborhood become the keys and values for the attention mechanism. The points outside are discarded. Consequently, the number of keys and values depends on the number of points present in the vicinity of the query point. 
    More details in \autoref{sec:model:temporal}. 
    \vspace{-0.5cm}
    }
    \label{fig:attention_temporal}
\end{figure}

To fuse the different timesteps, we introduce a \emph{Submanifold Attention} module employed in a temporal context, illustrated in \autoref{fig:attention_temporal} and inspired by window attention from the LiDAR 3D detection method \cite{lai2022stratified}. 
Formally, we represent a 2D BeV point $(x,y)$ at current time $t_0$ already processed by our model as a query $Q_{t_0,x,y}$.
Unlike more traditional approaches, the module is set with a spatiotemporal window $W := (w^t, w^x, w^y)$ where $w^t \in \mathbb{N}$ is a temporal range and $w^x, w^y \in \mathbb{N}$ are spatial windows 
{defining} a neighborhood around each query point.
For any given query point, only keys and values associated to points within the neighborhood of the query are considered when computing the attention.
{This new} attention mechanism can be written as:
\begin{equation*}
        O_{t_0,x,y} = \hspace{-0.2cm} \!\!\sum_{(t_k, x_k, y_k) \in W}\!\! \hspace{-0.2cm} A_{t_k,x_k,y_k} \Big(\frac{Q_{t_0,x,y} K_{t_k, x_k, y_k}^\top}{\sqrt{d_k}}\Big) V_{t_k, x_k, y_k}^\top,
\end{equation*}
where $A_{t_k,x_k,y_k}$ is the softmax term of the attention coefficients.
Due to the sparsity of the past data, each query may have varying numbers of keys and values.
This attention mechanism performs calculations only with point combinations within the specified window.
Note that an infinite spatiotemporal window recovers standard 
attention. 

\subsection{Inference with \acro{}}
\label{sec:model:inference}

To perform inference with \acro{} multiple strategies can be adopted.
First, for comparative evaluations against other approaches, a setting referred to as `dense {inference}' can be employed.
This involves processing all grid points with \acro{} in a single pass, producing predictions akin to considering the entire image, 
{as in a dense model.}

Alternatively and interestingly, we can also use sparse sampling \emph{during test time} to avoid allocating resources to regions of lesser interest.
This is similar to the `coarse/fine' training strategy with slight modifications.
We first need to define an exploration pattern for the coarse pass.
We can consider diverse options, including uniform sampling as used in training, a loose regular pattern arranged in a grid, or a distance-to-ego dependent pattern.
These strategies can be tailored based on specific use cases, such as emphasizing long-range sampling for highway driving or comprehensive, closer-range sampling for parking scenarios.
Besides, \acro{} can also leverage additional information present at test-time, when available.
For instance, LiDAR beams directly detect physical structures like roads or vehicles, providing valuable prior information for locating semantic classes.
Moreover, if available, we can use an HD map to minimize unnecessary computations in non-driveable regions such as buildings.

In the absence of such priors, we initialize the coarse pass by subsampling the BeV grid in evenly spaced locations, with each selected point spaced from its nearest points by a set parameter $k$ on the grid. $k=1$ recovers  {the dense grid}. %
{Then,} %
anchor points are selected where the model predicts a confidence above a %
threshold ${\tau \in [0,1]}$.
These anchor points are densified {with kernel $k_{fine} = k$} for the fine pass, mirroring the training strategy. 
{Coarser sub-sampling of the grid results in a larger $k_ {fine}$.}
Unless stated otherwise, this setting is used for our sparse regime evaluations.

\begin{figure}
    \centering
    \includegraphics[width=\columnwidth, trim=0cm 0cm 0cm 0cm, clip]{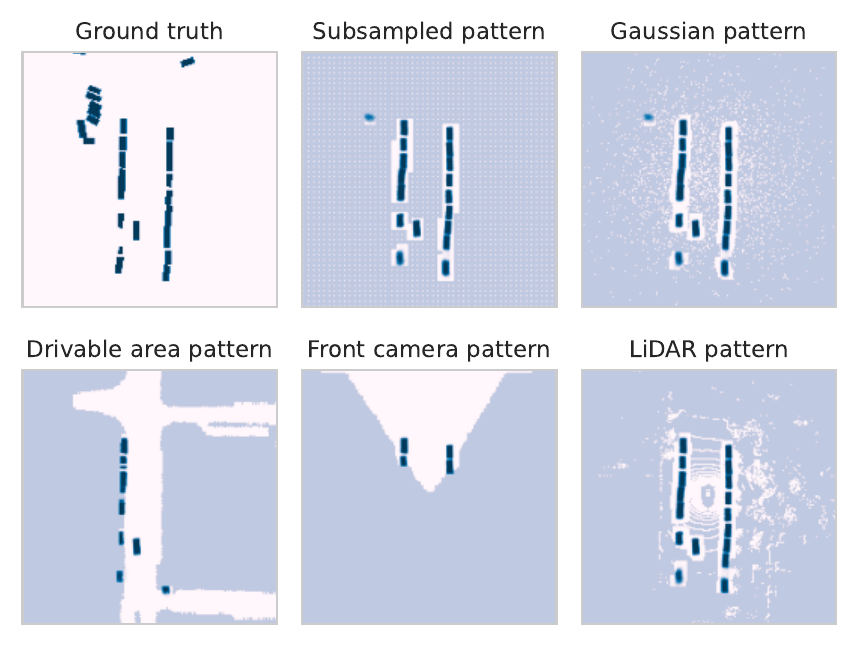}
    \caption{\textbf{Comparison of various sampling patterns with their predictions using the coarse-fine inference.} The inference coarse initialization can be declined in several strategies. On the first row we illustrated from left to right: the ground truth, the subsampled pattern and the Gaussian pattern concentrating more around points close to the ego-vehicle. On the second row we illustrate from left to right, the drivable area pattern, the front camera pattern and the LiDAR pattern. The predominantly pale colour indicates that no points have been sampled and that the associated prediction will be considered null.
    }
\label{fig:prediction_various_sampling}
\end{figure}

Using such methods, however, results in sparse predictions that do not cover the dense BeV.
To build a dense prediction map, for evaluation purposes typically, we simply consider the non-sampled locations as empty cells.
If ground-truth elements exist at these locations and are not predicted, they are counted as false negatives.
Illustrations of different strategies are shown in \autoref{fig:coarse_fine} and \autoref{fig:prediction_various_sampling}.

\section{Experiments}

\label{sec:exp}
\paragraph{Data, training and implementation details.}
\label{sec:exp:details}

Our experiments are conducted on nuScenes \cite{caesar2020nuscenes} {and Lyft L5 \cite{christy2019lyft} datasets}. NuScenes contains 1000 scenes split into 750-150-150 scenes for the training, validation, and test sets. {Lyft L5 contains 180 scenes, each 25-45 seconds in length, annotated at 5Hz,} {that we split as in FIERY \cite{fiery2021}.}
For all our experiments, \acro{} is trained on a maximum of 100 epochs %
using an Adam \cite{kingma2015adam} optimizer with learning rate $\lambda=3e^{-4}$, weight decay $w=10^{-7}$, a one-cycle linear learning rate scheduler. We used a 100m×100m grid with a 50cm resolution resulting in a 200x200 grid. For the training, unless specified, we used a random coarse strategy with $N_\textit{coarse} = N_\textit{fine} = 2.5k$ corresponding to 1/16 of the points of the grid. The densification patch size $k_\textit{fine}$ is set to 9. The number of anchor points is defined by $N_\textit{anchor} = 100$. We train our model using a segmentation, an offset and a centerness loss as in FIERY \cite{fiery2021}.
For sparse {inference,} %
the fine threshold is set to $\tau=0.1$. For the temporal models, the temporal threshold is set to $\tau_\textit{temp}=\text{sigm}(-5)$ and we use 2 seconds of context corresponding to 8 frames in the past. Details about the choice {and the robustness} %
of the parameters can be found in \autoref{sec:supp:sparse_inference}.
Unless stated otherwise, our experiments are done with a ResNet-50 \cite{he2016resnet} backbone, at image resolution $224 \times 480$, and without visibility filtering of low visilibity vehicle. 
An EfficientNet-b4 \cite{tan2020efficientnet} backbone can be introduced for fair comparisons with prior work.
The neck network is a simple upsampling with small convolutions between different resolutions, while the Sparse-UNet is coded using a sparse neural network library \cite{spconv2022} following the standard UNet architecture of Simple-BEV \cite{harley2022simple}. 
Our \textit{Submanifold Attention} module {is} coded using a graph neural network library \cite{fey2019torchgeom} and C++, and 
our efficient \textit{Sparse Feature Pulling} module implemented in CUDA . %

\subsection{State-of-the-art comparison}
\label{sec:exp:sota}

\begin{table}[t]
    \setlength{\tabcolsep}{.1cm}
    \centering
    \resizebox{\linewidth}{!}{
        \begin{tabular}{@{} l *{6}{@{\hspace{0.1cm}}c} @{}}
        \toprule
            \textbf{Vehicule segm. IoU ($\uparrow$)} \hspace{-3cm} & & & \multicolumn{2}{c}{\textit{No visibility filtering}}  & \multicolumn{2}{c}{\textit{Visibility filtering}} \\
            \cmidrule(lr){4-5}
            \cmidrule(lr){6-7}
             Method & Backb. & Temp. & $224\times480$ & $448\times800$ & $224\times480$ & $448\times800$ \\
        \midrule
            FIERY static \cite{fiery2021} & EN-b4 & & 35.8 & --- & 39.8 & --- \\ 
            CVT \cite{zhou2022crossview} & EN-b4 & & 31.4$^{*}$ & 32.5$^{*}$ & 36.0 & 37.7$^{*}$ \\ 
            LaRa \cite{bartoccioni2022lara}  & EN-b4 & & 35.4 & --- & 38.9 & --- \\ 
            BEVFormer \cite{li2022bevformer} & RN-50 & & 35.8$^{**}$ & 39.0$^{**}$ & 42.0$^{**}$ & 45.5$^{**}$ \\ 
            Simple-BEV \cite{harley2022simple} & RN-50 & & 36.9$^{*}$ & 40.9$^{*}$  & 43.0$^{*}$  & 46.6 \\ 
            BAEFormer \cite{Pan_2023_CVPR} & EN-b4 & & 36.0 & 37.8 & 38.9 & 41.0 \\ 
            \rowcolor{Apricot!20!white}
            \acro{} & EN-b4 &  & \textbf{38.7} & \textbf{42.1} & \textbf{44.0} & \textbf{47.6} \\
            \rowcolor{Apricot!20!white}
            \acro{} & RN-50 & & 38.1 & 41.7 & 43.7 & 47.0 \\
        \midrule
            FIERY \cite{fiery2021} & EN-b4 & \cmark & 38.2 & --- & --- & --- \\ 
            \rowcolor{Apricot!20!white}
            \acro{}-T & EN-b4 & \cmark & \textbf{39.9} & \textbf{43.2} & \textbf{44.7} & \textbf{48.7} \\  %
            \rowcolor{Apricot!20!white}
            \acro{}-T & RN-50 & \cmark & \textbf{39.9} & \textbf{43.2} & 44.1 & {47.7} \\       %
        \bottomrule
        \end{tabular}
    }
   \caption{\textbf{BeV vehicle segmentation on nuScenes.} computed on the validation set at different resolutions and for different filtering based on vehicle's visibility. No visibility filtering means all the annotated vehicles are considered. Visibility filtering means only the vehicles having a visibility $>40\%$ are considered. `*' indicates scores obtained using official codes, `**' indicates scores we obtained after reimplementing the model. In some cases, our reproduction yields higher scores than the ones reported in original papers (see Supplementary Materiel \ref{sec:supp:technical_details} for comparison details). `EN-b4' and `RN-50' stand for EfficientNet-b4 \cite{tan2020efficientnet} and ResNet-50 \cite{he2016resnet} respectively. `Temp' stands for `temporal' models using past frames. For fair comparisons we use 8 past frames.
   }
   \label{tab:sota:vehicle} 
\end{table}

\begin{table}[t]
    \setlength{\tabcolsep}{.1cm}
    \centering
    {
        \begin{tabular}{@{} l *{6}{@{\hspace{0.1cm}}c} @{}}
        \toprule
         \textbf{Vehicule segm. IoU ($\uparrow$)} & Long & Short \\
        \midrule
         FIERY & 36.7 & 59.4 \\
         BeVFormer (EN-b4) & 44.5 & 69.9 \\
         BeVFormer (RN-50) & 43.2 & 68.8 \\
         SimpleBEV (EN-b4) & 44.5 & 70.4 \\
         SimpleBEV (RN-50) & 43.6 & 70.7 \\
         \rowcolor{Apricot!20!white} {PointBeV} (EN-b4) & \textbf{45.4} & \textbf{72.6} \\
         \rowcolor{Apricot!20!white} {PointBeV} (RN-50) & \textbf{44.5} & \textbf{72.3} \\
         \bottomrule
        \end{tabular}
    }
   \caption{\textbf{BeV vehicle segmentation on Lyft L5.} Scores are IoU\,$(\uparrow)$, models are trained at $224 \times 480$ resolution for different backbones and {ranges: 30m $\times$ 30m (Short) and 100m $\times$ 100m (Long).} 
   Data splits are {from} %
   FIERY \cite{fiery2021}.
    \label{tab:sota:lyft}}
   \label{tab:sota:lyft} 
\end{table}

We compare the performance of \acro{} against a collection of BeV segmentation methods from the literature on vehicle, pedestrian, and lane segmentation tasks, in static and temporal settings, on nuScenes \cite{caesar2020nuscenes}.

For vehicle segmentation, \acro{} consistently outperforms existing methods, achieving state-of-the-art performance on nuScenes \cite{caesar2020nuscenes} {and Lyft L5 \cite{christy2019lyft}} across various settings (\autoref{tab:sota:vehicle}, \autoref{tab:sota:lyft}) . Our method exhibits superiority at different resolutions ($224 \times 480$ or $448 \times 800$), under different standard visibility settings, for different backbones, and for both static and temporal settings.
In the absence of reported results in original papers, we either run the original codes or we reimplement the method.
{We ensure reproductions are consistent with the original scores in \autoref{sec:supp:technical_details}.}

We expand our evaluations to pedestrian %
(\autoref{tab:sota:pedestrian}) and lane (\autoref{tab:sota:lane}) segmentation.
Remarkably, without adjusting the number of training points nor the patch sizes, our approach sets new state-of-the-art %
for these tasks.
{\acro{} surpasses previous state-of-the-art by +1.3 IoU points for pedestrians (TBP-Former \cite{fang2023tbpformer}) in both static and temporal regimes, and by +4.8 IoU points for lanes (MatrixVT \cite{zhou2022matrixvt}).}

\begin{table}[t]
        \centering
            \begin{tabular}{@{}l @{\hspace{0.1cm}} c @{\hspace{0.1cm}} c@{}}
            \toprule
                \textbf{Pedestrian segm.} & \textbf{Temp.} & \textbf{IoU} ($\uparrow$) \\
            \midrule
                LSS \cite{philion2020lift} & & 15.0  \\
                FIERY \cite{fiery2021} & & 17.2  \\ 
                ST-P3 \cite{hu2022stp3} & & 14.5  \\ 
                TBP-Former static\,\cite{fang2023tbpformer} & & 17.2 \\ 
                \rowcolor{Apricot!20!white}
                \acro{} & & \textbf{18.5} \\ 
                \midrule
                TBP-Former \cite{fang2023tbpformer} & \cmark & 18.6  \\ 
                \rowcolor{Apricot!20!white}
                \acro{}-T & \cmark & {19.9} \\ 
            \bottomrule
            \end{tabular}
            \caption{\textbf{BeV pedestrian segmentation on nuScenes.} Scores are IoU\,$(\uparrow)$ with visibility filtering, computed on the validation set at $224 \times 480$ resolution. `Temp.' refers to temporal models using 8 contextual frames for fair comparisons.
            \label{tab:sota:pedestrian}}
\end{table}

\begin{table}
    \centering
        \begin{tabular}{@{} l c @{}}
        \toprule
            \textbf{Lane segm.} & \textbf{IoU} ($\uparrow$) \\
        \midrule
            BEVFormer \cite{li2022bevformer} & 25.7 \\
            PETRv2 \cite{liu2022petrv2} & 44.8 \\
            M$^2$BEV \cite{xie2022m2bev} & 38.0 \\
            MatrixVT \cite{zhou2022matrixvt} & 44.8 \\
            \rowcolor{Apricot!20!white}
            \acro{} & \textbf{49.6} \\
        \bottomrule
        \end{tabular}
        \caption{\textbf{BeV lane segmentation on nuScenes \cite{caesar2020nuscenes} validation set.} Scores are IoU\,${(\uparrow)}$, computed using models trained at $224 \times 480$ image resolution. All models are static. \label{tab:sota:lane}}
\end{table}

\subsection{Ablations}
\label{sec:exp:ablations}

To validate the %
different components of our model, we conduct an ablation study focusing on the efficiency and impact of key modules, including our custom \textit{Sparse Feature Pulling} module (\autoref{sec:model:sparse_feature}), our \textit{Submanifold Attention} module (\autoref{sec:model:temporal}), and the coarse/fine {training passes} %
(\autoref{sec:model:coarse_fine}).

\vspace{.2cm}\noindent\textbf{Sparse feature pulling.}
To assess the efficiency of our \textit{Sparse Feature Pulling} module, we conduct a comparative evaluation of execution time and memory usage between the default module and our proposal. The results are summarized in \autoref{tab:ablation:fps_memory_per_bs}.
They reveal a notable improvement in memory efficiency, ranging between 25\% to 45\%, when utilizing the sparse interpolation module. Simultaneously, there is a marginal increase in FPS while GPU modules are not fully optimised to handle a different number of keys and values.
Interestingly, the use of the module enables accommodating larger batches, notably up to a batch size of 28 on a 40GiB A100 GPU. This enhancement is key, enabling faster training with larger batches, the consideration of extended temporal contexts, and supporting increased BeV range and resolution.
For detailed insights into the module's specific performance, we provide further investigations in  \autoref{sec:supp:sparse_feature_pulling}. %

\begin{table}[t]
    \centering
    \resizebox{\linewidth}{!}{
    \begin{tabular}{@{}l *{8}{@{\hspace{2mm}}c} @{}} 
    \toprule
    & \multicolumn{2}{c}{bs=1} & \multicolumn{2}{c}{bs=2} & \multicolumn{2}{c}{bs=12} & \multicolumn{2}{c}{bs=28} \\
    \cmidrule(lr){2-3}
    \cmidrule(lr){4-5}
    \cmidrule(lr){6-7}
    \cmidrule(lr){8-9}
    & Mem & FPS & Mem & FPS & Mem & FPS & Mem & FPS \\
    \hline
    \rowcolor{Apricot!20!white}
     \acro{} & 3.50 & 14.84 & 5.50 & 10.01 & 25.99 & 2.30 & 39.40 & 1.00 \\
     \hspace{0.2cm} w/o Sparse Feature Pulling & 4.57 & 13.70 & 7.61 & 9.81 & 38.94 & 2.12 & --- & --- \\ 
    \bottomrule
    \end{tabular}
    }
    \caption{\textbf{Ablation of the Sparse Feature Pulling module.} 
    Memory consumption and FPS for our model using or not our sparse module, computed on
    a 40GB A100 under dense {inference} %
    {setting}. `---' indicates `out of memory'. `Mem' is for Memory footprint in GiB, and `bs' stands for batch size.}
    \label{tab:ablation:fps_memory_per_bs}
\end{table}

\vspace{.2cm}\noindent\textbf{Submanifold Temporal Attention.}
\begin{table}[t]
    \setlength{\tabcolsep}{.1cm}
    \centering
    \resizebox{\linewidth}{!}{
        \begin{tabular}{@{} l *{6}{@{\hspace{0.1cm}}c} @{}}
        \toprule
            \textbf{Vehicle segm.\ IoU ($\uparrow$)} \hspace{-4cm} & & & \multicolumn{2}{c}{\textit{No visibility filtering}}  & \multicolumn{2}{c}{\textit{Visibility filtering}} \\
            \cmidrule(lr){4-5}
            \cmidrule(lr){6-7}
            Method & Temp. & $N_\textit{point}$ & $224\times480$ & $448\times800$ & $224\times480$ & $448\times800$ \\
        \midrule
            \acro{} & & 40k & 38.09 & {41.66} & 43.70 & {47.20} \\
            \midrule
            \acro{}-T & \cmark & $\sim$66k & 39.93 & 43.19 & 44.06 & {47.67} \\
            \hspace{0.2cm} w/o subman. att. & \cmark & 320k & 40.49 & {43.53} & {44.12} & 47.63 \\
        \bottomrule
        \end{tabular}
    }
   \caption{\textbf{Ablation of the submanifold temporal attention.} Scores are IoU ($\uparrow$) computed on the nuScenes \cite{caesar2020nuscenes} validation set at different resolutions and 
   {vehicle's visibility filterings}.
   }
   \label{tab:ablation:temporal_dense_sparse} 
\end{table}

In \autoref{tab:ablation:temporal_dense_sparse}, we conduct an ablation study focusing on the Submanifold Temporal attention. Although the results can be very marginally improved with standard attention in some settings, achieving this enhancement comes with the cost of considering approximately $\sim$5$\times$ more points and computations.
Leveraging the submanifold temporal attention unlocks new possibilities for considering longer temporal contexts.
Accordingly, we trained and evaluated \acro{} under higher number of frames, up to 25, and obtained an IoU of 40.73.

\vspace{.2cm}\noindent\textbf{Coarse and fine training.}
\autoref{tab:ablation:fine_pass} displays the performance evolution of BeV vehicle segmentation across various point budgets \emph{during training}, with similar training times for all models. Remarkably, \acro{} achieves similar performances with only 20\% of the total points %
{instead of the whole}
BeV grid.
{It shows how}
the use of sparse training approach in \acro{} 
allows to optimize memory usage effectively, e.g., to train on longer temporal contexts.
\begin{table}[t]
    \centering
    \resizebox{\linewidth}{!}{
    \begin{tabular}{@{} l *{4}{c} @{}}
    \toprule
    $N_\textit{point}$ (\% of full BeV) & 4k (10\%) & 8k (20\%) & 20k (50\%) & 40k (100\%) \\
    \midrule
     \acro{} & 38.1 & 38.3 & 38.4 & 38.3 \\
    \bottomrule
    \end{tabular}
    }
    \vspace{0.15cm}
    \caption{Evolution of BeV vehicle segmentation scores \textbf{IoU ($\uparrow$) for various point budgets.} The budget is given as total number of points used in both passes and expressed as a percentage of the total number of points in a dense grid.
    With only 20\% of points, we find similar performances as with 100\%. %
    }
    \label{tab:ablation:fine_pass}
\end{table}

Additionally, to validate the efficiency of our two-stage training strategy involving coarse and fine passes, we analyze two settings based on a given point budget $N_\textit{point}$. The first setting uses only a coarse pass (${N_\textit{coarse} = N_\textit{point}}$, $N_\textit{fine} = 0$) and achieves an IoU of 35.4. In contrast, the second setting is our coarse/fine strategy and divides the point budget equally between both passes (${N_\textit{coarse} = N_\textit{fine} = N_\textit{point} / 2}$), yielding an IoU of 38.3 (as depicted in \autoref{tab:ablation:fine_pass}).
These results highlight the superiority of splitting points between sequential coarse and fine passes. Besides, we find that the two-pass approach considerably stabilizes training.

\subsection{Adaptive Inference Capabilities}

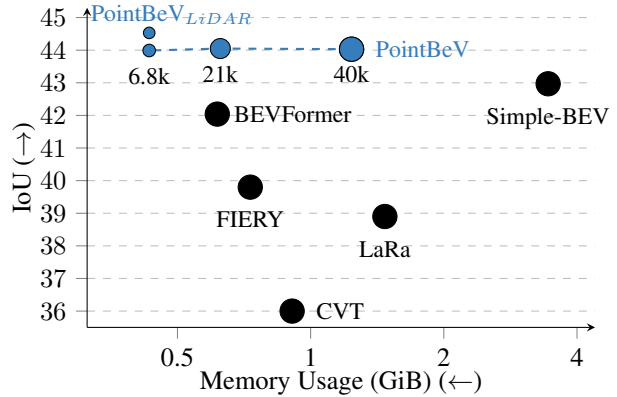
\begin{figure}[t]
\begin{tikzpicture}
\centering %
\begin{axis}[
    width=\linewidth, %
    height=.7\linewidth,
    xlabel={Memory Usage (GiB) ($\leftarrow$)},
    ylabel={IoU ($\rightarrow$)},
    xmin=320, xmax=4500,
    ymin=35.5, ymax=45.3,
    xmode=log, %
    log basis x={2}, %
    xtick={512,1024,2048,4096},
    xticklabels={0.5,1,2,4},
    ytick={36, 37, 38, 39, 40, 41, 42, 43, 44, 45},
    xlabel style={yshift=5pt}, %
    ylabel style={yshift=-5pt}, %
    scaled ticks=false, %
    ymajorgrids=true,
    grid style=dashed,
    axis lines=left, %
    tick align=outside,
    nodes near coords align={vertical},
    every node near coord/.append style={
        font=\small,  %
    }
]

\def\basePointSize{2pt} %

\newcommand{\scaledPoint}[6]{%
    \pgfmathparse{log10(#1)/log10(5)} %
    \edef\pointSize{\pgfmathresult*\basePointSize}
    \edef\temp{\noexpand\addplot[
        point plot style,
        mark options={fill=#4, scale=\pgfmathresult},
        every node near coord/.append style={#6},
    ] coordinates {
        (#2,#3) [#5]
    };}
    \temp
}

\scaledPoint{40}{1267}{44.03}{cvprblue}{\textcolor{cvprblue}{PointBeV}}{anchor=west, xshift=2mm}
\scaledPoint{40}{1267}{44.03}{cvprblue}{40k}{anchor=north, yshift=-1mm}
\scaledPoint{21}{640}{44.05}{cvprblue}{21k}{anchor=north,yshift=-1mm}
\scaledPoint{6.8}{442}{43.99}{cvprblue}{6.8k}{anchor=north,yshift=-1mm}

\addplot[
    color=cvprblue,
    dashed,
    thick
] 
coordinates {
    (1267,44.03)
    (640,44.05)
    (442,43.99)
};

\scaledPoint{5.8}{442}{44.53}{cvprblue}{\textcolor{cvprblue}{PointBeV$_{LiDAR}$}}{anchor=south,xshift=3mm}

\scaledPoint{40}{3525}{42.97}{black}{Simple-BEV}{anchor=north, yshift=-2mm}
\scaledPoint{40}{630}{42.04}{black}{BEVFormer}{anchor=west, xshift=1mm}
\scaledPoint{40}{930}{36.0}{black}{CVT}{anchor=west, xshift=2mm}
\scaledPoint{40}{1507}{38.9}{black}{LaRa}{anchor=north, yshift=-2mm}
\scaledPoint{40}{748}{39.8}{black}{FIERY}{anchor=north, yshift=-2mm}

\end{axis}
\end{tikzpicture}

\caption{\textbf{BeV vehicle IoU vs. memory footprint.}
The size of a dot represents the number of BeV points being evaluated (the smaller the better).
\acro{} has the capacity to explore various trade-offs between efficiency (Memory usage) and the performances (IoU) by varying the number of points being evaluated. Here, we show PointBeV models with an EfficientNet-b4, and vehicle with low visibility are filtered out (unlike in \autoref{fig:iou_vs_mem}).
}
\label{fig:iou_vs_mem_vis2}
\end{figure}

One intriguing aspect of \acro{} lies in its inherent flexibility to adjust between efficiency and accuracy, and to adapt to the demands of specific use cases.
This is illustrated in \autoref{fig:iou_vs_mem} and \autoref{fig:iou_vs_mem_vis2}, where the choice of the number of points directly influences both performance and memory footprint.
Interestingly, achieving a comparable IoU score (approximately 38.3 for \autoref{fig:iou_vs_mem} and 44.0 for \autoref{fig:iou_vs_mem_vis2}) is feasible by employing only one-sixth of the total points, significantly reducing memory usage by a third at inference time.
This underscores the model's efficient utilization of computational resources without compromising performance.
{Note that our general sparse sampling setting (\autoref{sec:model:inference}) is chosen to perform well in most common scenarios. For particular cases such as small or very large objects, we anticipate that it can be further improved with specific strategies.
}

Lastly, as depicted in \autoref{fig:prediction_various_sampling} and previously discussed in \autoref{sec:model:inference}, \acro{} exhibits the capability to integrate external priors to guide its focus.
These priors can stem from various sources such as LiDAR point clouds, which reveal physical structures, or HD maps that aid in reducing sampling in regions irrelevant to the task, such as looking for agents within buildings.
In \autoref{fig:iou_vs_mem_vis2}, we see that by sampling points with this LiDAR prior, superior IoU can be achieved compared to the dense setting (\acro{}$_\textit{LiDAR}$ reaches 44.5 vs 44.0 for default sampling of \acro{}), with a much lower number of points and memory usage. By leveraging LiDAR data, \acro{}$_\textit{LiDAR}$ minimizes noise in empty regions leading to improved segmentation accuracy.
Otherwise, a coarse pass based on a Gaussian prior around the ego car could be used to reinforce attention to closer ranges.

\section{Conclusion}

\label{sec:conclusion}

We introduced \acro{} for BeV segmentation from camera inputs. By integrating sparse modules and an innovative training strategy, our method operates efficiently while setting a new state-of-the-art with or without temporal information, on nuScenes segmentation benchmarks. 
\acro{} also accommodates to additional test-time information or to low compute regimes without retraining.

This work is only a first step in exploring the potential of the sparse paradigm for BeV segmentation.
Upcoming directions might encompass 2D occupancy forecasting and 3D voxel occupancy estimation applications. These tasks would immediately benefit from the reduced memory footprint for longer temporal horizons or higher spatial resolutions.
Our sparse inference regime uses a simple dense BeV completion by setting every non-selected point as empty; introducing a light completion network may yield additional 
gains. 
Further down the road, 
future directions may explore dynamic sampling strategies guided by sensor inputs, by following agent flows to determine sampling locations for subsequent frames, or by the need of downstream tasks such as forecasting or planning.
Finally, this new sparse paradigm is also very suited to the unification of BeV tasks at diverse resolutions and ranges within the same model. 

\vspace{.5cm}
\noindent\textbf{Acknowledgments.} This paper is dedicated to Laura E-R.
We received support of the French Agence Nationale de la Recherche (ANR), under grant ANR-21-CE23-0032 (project MultiTrans).
This work was performed using HPC resources from GENCI–IDRIS (Grant 2023-AD011014252).
In addition we thank Yihong Xu and Alexandre Boulch for their contributions and exchanges of ideas throughout the process, particularly with regard to the implementation of the submanifold temporal attention.

{\small
\bibliographystyle{ieeenat_fullname}
\bibliography{egbib}
}

\clearpage
\setcounter{page}{1}
\appendix

\maketitlesupplementary
\renewcommand{\appendixautorefname}{Section} %

This document contains technical details about training and implementation of models (\autoref{sec:supp:technical_details}), further memory consumption studies in (\autoref{sec:supp:memory}), and additional details on our Sparse Feature Pulling module (\autoref{sec:supp:sparse_feature_pulling}) and on our temporal model (\autoref{sec:supp:temporal_model}).
We also detail more the sparse inference settings (\autoref{sec:supp:sparse_inference}). Finally we display some predictions (\autoref{sec:supp:qualitative}).

\section{Technical Details}
\label{sec:supp:technical_details}

\subsection{\acro{} Training}
We train \acro{} using both image \cite{harley2022simple} and BeV augmentations \cite{zhang2023sabev}.
For BeV augmentations, viewpoint changes (translations and rotations) are encoded in an augmentation matrix.
This matrix is applied on the bounding box coordinates before building the ground-truth segmentation map, and when projecting the pillar points points in camera coordinates in the forward pass.
We train static models for a maximum of 100 epochs, as they always converge earlier.

{The number of training epochs is an upper bound, we get similar performance earlier ($\sim$50 epochs in low resolution with filtering, and $\sim$80 for high resolution without filtering).
Moreover, \acro{} trained for 30 epochs already performs beyond previous SOTA (see table below),  and each epoch is also much faster (15min/epoch for \acro{} vs.\ 52min/epoch for BEVFormer on a single 40GB A100 GPU with maximum model batch size). 
\begin{table}[h]
    \centering
    \resizebox{\linewidth}{!}{
    \begin{tabular}{@{} l *{4}{@{\hspace{0.2cm}}c} @{}}
        \toprule
        \textbf{IoU ($\uparrow$) vehicle} & \multicolumn{2}{c}{\textit{No vis.\ filtering}}  & \multicolumn{2}{c}{\textit{Vis.\ filtering}} \\
        Backbone: EN-b4 & $224\!\times\!480$ & $400\!\times\!800$ & $224\!\times\!480$ & $400\!\times\!800$ \\
        \midrule
         PointBeV @30epochs& 38.1 & 41.6 & 43.1 & 46.8 \\
         \bottomrule
    \end{tabular}
    }
\end{table}
}

For the temporal model, \acro{}-T, we start from a static checkpoint and we add a single submanifold temporal attention layer.
The temporal layer is trained using 8 past frames for fair comparisons corresponding to 2 seconds while the rest of the network is kept frozen.

\subsection{Comparison with the Baselines}
In \autoref{tab:sota:vehicle}, we take the official numbers from the papers whenever available.
However, across different publications, the numbers are often reported in different settings in terms of image resolution or vehicle filtering, preventing direct comparison.
Therefore, to provide a more complete view, we trained the models on the 4 different settings on which we compare them.
We stress that the models are re-trained specifically for these settings, in opposition to simply evaluating the official checkpoints on different settings.

In cases where the code is available, such as for CVT \cite{zhou2022crossview} and Simple-BEV \cite{harley2022simple}, we use it and only change the image resolution and the visibility filtering.
The official CVT \cite{zhou2022crossview} repository already includes a metric that accounts for visibility so we simply modify the visibility applied in the loss function before retraining the models.
For Simple-BEV \cite{harley2022simple}, we modify the code, particularly the dataloader, to incorporate the visibility annotations.
To ensure that the implementations are correct, we compare in \autoref{model_comparison_paper_repro} the results of our reproduction against those reported in the papers, in their proposed settings.
We see that they are very similar.

When the code is not available, we replicated the method. This is the case for BEVFormer \cite{li2022bevformer}, which does not have an official segmentation code.
For BEVFormer, we used 6 layers defined by one deformable self-attention followed by one deformable cross-attention with 4 heads and 8 offsets per points.
We are comparing \acro{} with the static BEVFormer model, and for fair comparisons, we are using single-scale image features as indicated in the reproduction code of the official Simple-BEV repository. With BEVFormer, we achieved results superior  to those reported in the original paper by utilizing a lower image resolution, which validates the reproduction code (see \autoref{model_comparison_paper_repro}).

\begin{table}[h]
    \centering
\resizebox{\linewidth}{!}{%
\setlength{\tabcolsep}{2pt}

\begin{tabular}{@{} l @{} ccc c @{}}
\toprule
\textbf{Method} & Resolution & Visibility & IoU orig. & IoU reproduced \\ \midrule
Simple-BEV \cite{harley2022simple} & 448$\times$800 & with filtering & 46.6 & 46.56 \\
CVT \cite{zhou2022crossview} & 224$\times$480 & with filtering & 36.0 & 36.63 \\
\multirow{2}{*}{BEVFormer \cite{li2022bevformer}}  &  640$\times$1600 & with filtering & 44.4 & --- \\
& 448$\times$800 & with filtering & --- & 45.56 \\ \bottomrule
\end{tabular}
}
    \caption{\textbf{Comparison of the results obtained after training with the official code} under the setting of the paper and those reported in the papers. Our experiments reached similar results.}
    \label{model_comparison_paper_repro}
\end{table}

\subsection{Parameter Count}

We analyze the number of parameters for various models considered. The parameter counts were obtained from the official codes. For a fair comparison we report the number of parameters using the same ResNet-50 \cite{he2016resnet} backbone which also influences the neck network when there is one. The neck network of \acro{} considers two resolutions, returned by the backbone, and aligns them by applying bilinear interpolation to the smaller one. The channels of these resolutions are then concatenated, followed by a series of three convolutions to adjust the final channel dimension. 

\begin{table}[h]
\centering
\resizebox{\linewidth}{!}{%
\setlength{\tabcolsep}{2pt}
\begin{tabular}{@{} l @{} cccccc | c @{}}
\toprule%
\textbf{Method} & Backbone & Neck & VT & Update & Temporal & Heads & Total\\
\midrule
CVT \cite{zhou2022crossview} & 8.5M & --- & 819k & 244k & --- & 37.1k & 9.6M  \\
LaRa \cite{bartoccioni2022lara}& 8.5M & 1.9M & 2.6M & 4.9M & --- & 295k & 18.2M \\
BEVFormer \cite{li2022bevformer}& 8.5M & 9.5M & 7.3M & --- & --- & 442k & 25.8M  \\
\rowcolor{Apricot!20!white}
PointBEV & 8.5M & 9.5M & 291k & 3.6M & 564k & 442k & 22.9M \\
\bottomrule
\end{tabular}%
}
\caption{\textbf{Comparison of the number of parameters of several models.} For a fair comparison we used the same backbone, i.e., ResNet-50 \cite{he2016resnet}. The column `VT' corresponds to `View Transform', that is any learnable operation involved in the camera-to-BEV projection (e.g., a cross-attention for CVT and LaRa \cite{bartoccioni2022lara}, all deformable blocks for BEVFormer \cite{li2022bevformer}). If available, BeV learnable grid parameters are accounted in the view transform column.}
\label{tab:models_params}
\end{table}

\section{Further memory analysis}
\label{sec:supp:memory}
To complement the memory analysis described in the main paper (\autoref{fig:iou_vs_mem}, \autoref{fig:iou_vs_mem_vis2}), we estimate the maximum training and validation batch sizes of various models on a 40GB A100 by studying the out-of-memory boundary in several forwards and backwards.
The aim is to study how the models scale and at what point they reach memory saturation.
During validation, the results in \autoref{tab:mem_valid_batch_size_comparison} indicate that PointBeV scales better than other models using an EfficientNet-b4 \cite{tan2020efficientnet}, achieving a $2\times$ increase in maximum validation batch size compared to LaRa \cite{bartoccioni2022lara} and a $4\times$ increase compared to CVT \cite{zhou2022crossview}. Therefore, the model is more suited for high-resolution tasks than the preceding models. 

\begin{table}[t]
\resizebox{\linewidth}{!}{%
\setlength{\tabcolsep}{2pt}
\begin{tabular}{lcccc}
\toprule
\multirow{2}{*}{\textbf{Method}} & \multirow{2}{*}{\textbf{Backbone}} & \multicolumn{3}{c}{\textbf{Resolution}} \\ \cmidrule(lr){3-5} 
 &  & \multicolumn{1}{c}{$224 \times 480$} & \multicolumn{1}{c}{$448 \times 800$} & \multicolumn{1}{c}{$640 \times 1600$} \\ \midrule
LaRa \cite{bartoccioni2022lara} & EN-b4 & \multicolumn{1}{c}{27} & \multicolumn{1}{c}{17} & 5 \\
CVT \cite{zhou2022crossview} & EN-b4 & \multicolumn{1}{c}{38} & \multicolumn{1}{c}{12} & 3 \\
Simple-BEV \cite{harley2022simple} & RN-50 & \multicolumn{1}{c}{11} & \multicolumn{1}{c}{11} & 10 \\
BEVFormer \cite{li2022bevformer} & RN-50 & \multicolumn{1}{c}{71} & \multicolumn{1}{c}{63} & 23 \\
\rowcolor{Apricot!20!white}
PointBeV & EN-b4 & \multicolumn{1}{c}{31} & \multicolumn{1}{c}{30} & 11 \\
\rowcolor{Apricot!20!white}
PointBeV & RN-50 & \multicolumn{1}{c}{31} & \multicolumn{1}{c}{28} & 17 \\ \bottomrule
\end{tabular}
}
\caption{\textbf{Memory analysis of various models by comparing maximum batch sizes during \emph{inference}} on a 40GB A100. `EN-b4' refers to EfficientNet-b4 \cite{tan2020efficientnet}, and `RN-50' to ResNet-50 \cite{he2016resnet}.}
\label{tab:mem_valid_batch_size_comparison}
\end{table}

For training, as seen in \autoref{tab:mem_train_batch_size_comparison}, PointBeV can handle significantly larger batch sizes and scales better in terms of resolution compared to all other models.

\begin{table}[t]
\resizebox{\linewidth}{!}{%
\setlength{\tabcolsep}{2pt}
\begin{tabular}{lcccc}
\toprule
\multirow{2}{*}{\textbf{Method}} & \multirow{2}{*}{\textbf{Backbone}} & \multicolumn{3}{c}{\textbf{Resolution}} \\
\cmidrule(lr){3-5}
 &  & \multicolumn{1}{c}{$224 \times 480$} & \multicolumn{1}{c}{$448 \times 800$} & \multicolumn{1}{c}{$640 \times 1600$} \\ \midrule
LaRa \cite{bartoccioni2022lara} & EN-b4 & 5 & 3 & 1 \\
CVT \cite{zhou2022crossview} & EN-b4 & 5* & 1* & 1* \\
Simple-BEV \cite{harley2022simple} & RN-50 & 8 & 5 & 2 \\
BEVFormer \cite{li2022bevformer} & RN-50 & 10 & 8 & 3 \\
\rowcolor{Apricot!20!white}
PointBeV & EN-b4 & 13 & 4 & 1 \\ 
\rowcolor{Apricot!20!white}
PointBeV & RN-50 & 26 & 9 & 3 \\ \bottomrule
\end{tabular}
}
\caption{\textbf{Memory analysis of various models by comparing maximum batch sizes during \emph{training}} on a 40GB A100. `EN-b4' refers to EfficientNet-b4 \cite{tan2020efficientnet}, and `RN-50' to ResNet-50 \cite{he2016resnet}. `*' indicates that the model has been trained without checkpointing at the backbone level for a fairer comparison between models. Checkpointing has a direct influence on the memory footprint.}
\label{tab:mem_train_batch_size_comparison}
\end{table}

\section{Sparse Feature Pulling}
\label{sec:supp:sparse_feature_pulling}

In the context of BeV projection, the feature pulling module takes as input a grid of image features and a list of coordinates.
It outputs a list of image features, corresponding to the provided coordinates.
The problem is that existing interpolation modules only work with a fixed number of points per batch and per camera.
This enforces models to consider more points than the number of visible points per camera.
We propose a custom interpolation module that removes this limitation by introducing a batch reference table.
\autoref{tab:ablation:grid_sampling} shows the benefits in terms of speed and memory of our module in standard conditions for BeV methods.
We compared the native and custom modules under the standard use case. 
In details, we consider a 3D pillar BeV of $X \times Y \times Z = 200 \times 200 \times 8$ points, and 6 feature images, one per camera. This setting is the one we encounter in nuScenes \cite{caesar2020nuscenes}.
Each feature image has 128 channels and their resolution is the down-sampled one after the backbone, i.e., a $224 \times 480$ original resolution leads to a down-sampled $28 \times 60$ resolution. 
Compared to its torch-based counterpart (\autoref{tab:ablation:grid_sampling}), our sparse interpolation module does not calculate the features of points not visible in the cameras, resulting in faster computation and a smaller memory footprint. Note that conventional use of the torch module requires to apply after the interpolation a masking operation to remove unused features. We took this operation into account in our table. 
Our module demonstrates significantly lower memory and time usage. It is 2.3$\times$ faster for forward and 4.2$\times$ for backward, with over 2$\times$ less memory consumption in both cases.

\begin{table}[t]
    \centering
    \resizebox{\linewidth}{!}{
    \begin{tabular}{@{}l @{\hspace{2mm}} c @{\hspace{2mm}} c @{\hspace{2mm}} c @{\hspace{2mm}} c @{}}
    \toprule
     & \multicolumn{2}{c}{\textit{Forward}} & \multicolumn{2}{c}{\textit{Backward}} \\
     \cmidrule(lr){2-3}
     \cmidrule(lr){4-5}
    \textbf{Module} & Mem (GiB) & Time (ms) & Mem (GiB) & Time (ms) \\
    \midrule
     Naive Feature Pulling & 1.9 & 4.3 & 2.8 & 26.0 \\ 
     \rowcolor{Apricot!20!white}
     Sparse Feature Pulling & 0.9 & 1.9 & 1.4 & 6.2 \\
    \bottomrule
    \end{tabular}
    }
    \caption{\textbf{Sparse Feature Pulling module memory and time footprints.} Results are for a batch composed of a single example.}
    \label{tab:ablation:grid_sampling}
\end{table}

\section{Temporal model}
\label{sec:supp:temporal_model}

One goal of the submanifold temporal attention module is to apply attention to a reduced combination of points. Therefore, we established a threshold $\tau_\textit{temp}$ at which a point is considered temporally active or inactive. To asses the number of points filtered at each time step, we used a static model and analyzed the distribution of logits in the prediction map (\autoref{tab:temp_threshold_active_pts}). To be conservative, we set our threshold at the inflection point of the static model, i.e., at $\tau_\textit{temp}=\text{sigm}(-5)$. Given the threshold considered, our module processes on average only one tenth of the points in the past, which divides the calculations by 10 compared with naive temporal attention in torch.

\begin{table}[h]
    \centering
\resizebox{\linewidth}{!}{%
\setlength{\tabcolsep}{2pt}
\begin{tabular}{cccccccc}
\toprule
\textbf{$\tau_\textit{temp}$} & $\text{sigm}(0)$ & $\text{sigm}(-3)$ & $\text{sigm}(-5)$ & $\text{sigm}(-6)$ & $\text{sigm}(-7)$ & $\text{sigm}(-8)$ & $\text{sigm}(-9)$  \\ 
\midrule
\textbf{\# Points} & 725 & 1932 & 3895 & 6091 & 11266 & 27084 & 38442 \\ 
\textbf{IoU } & 37.0 & 39.1 & 39.9 & 40.0 & 40.2 & 40.2 & 40.2 \\ 
\bottomrule
\end{tabular}
}
\caption{\textbf{Analysis of the number of activated points according to the applied temporal threshold $\tau_\textit{temp}$}. To do this, we take a static model trained at $224 \times 480$ image resolution without visibility filtering having 39.9 IoU and calculated how many points were above the considered threshold. The corresponding IoU is the temporal model evaluated using the temporal threshold.}
\label{tab:temp_threshold_active_pts}
\end{table}

\section{Sparse Inference}
\label{sec:supp:sparse_inference}

The introduction of sparse inference highlighted several parameters, such as the size of the densification patch and the threshold for considering a coarse point as an anchor point. To thoroughly analyze the impact of these metrics on sparse evaluation, we retrieved the checkpoint of a model trained without the visibility filter at a resolution of 224$\times$480 with an associated 38.09 IoU, and then we varied the aforementioned parameters during inference. By default, the number of coarse points associated with its model is $N_\textit{coarse}=2500$, the threshold $\tau=0.1$, the fine patch size is $k_\textit{fine}=9$.

\subsection{Fine patch size}

When varying the size of the densification patch, we observe that smaller patches lead to lower results, mainly due to having only 1/16 of the total points in the coarse pass (\autoref{tab:ablation_patch_size}). However, considering the geometric pattern of the coarse pass, the IoU plateaus beyond a certain window size. This is logical, as the spacing between two coarse points becomes less than half of the window size.

\begin{table}[h]
    \centering
    \resizebox{\linewidth}{!}{
    \setlength{\tabcolsep}{3pt}
    \begin{tabular}{@{}l *{7}{c} @{}}
    \toprule
    \textbf{Patch size} & 1 & 3 & 5 & 7 & 9 & 11 & 13 \\
    $N_\textit{fine}$ & 92 & 826  & 1,743 & 2,277 & 2,839 & 3,414 & 4,009 \\
    Memory (MB) & 461 & 462 & 472 & 481 & 493 &  507 &  522 \\
    \midrule
     \textbf{IoU} vehicle ($\uparrow$) & 2.6  & 22.0 & 37.3 & 38.0 & 38.1 & 38.1 & 38.1 \\  
    \bottomrule
    \end{tabular}
    }
        \caption{\textbf{Sparse evaluation on the nuScenes \cite{caesar2020nuscenes} validation set} of our model at resolution $224 \times 480$ without visibility filtering, using different patch sizes for the fine pass.
        Coarse sampling is a regular $50 \times 50$ grid (2,500 points). The reported memory is the maximum memory allocated calculated without the backbone.}
        \label{tab:ablation_patch_size}
\end{table}

Qualitatively, the larger the patch size $k_\textit{fine}$, the greater the number of activated points in the fine pass. Beyond a certain point, when half the size of the patch exceeds the distance between two neighboring points in the first pass, we have a prediction that nearly does not change anymore.
Considering the trade-off between the number of points considered, the patch size, and the final IoU, we have thus selected a default patch size of $k_\textit{fine}=9$, resulting in an IoU of 38.09 as indicated in \autoref{tab:sota:vehicle}.

\begin{figure*}
    \centering
    \includegraphics[width=\textwidth, trim=12cm 0 11cm 0, clip]{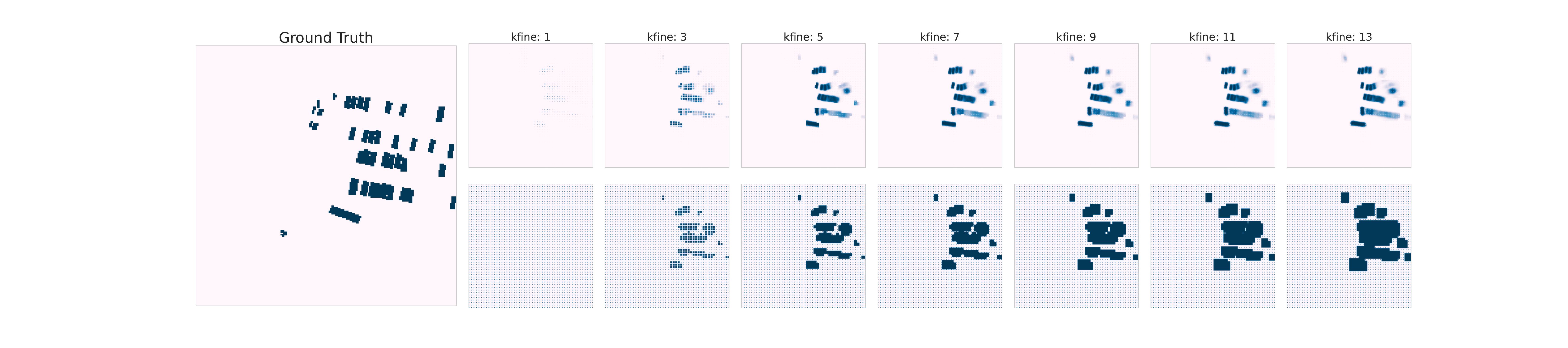}
    \caption{\textbf{Qualitative representation of different fine patch size $k_\textit{fine}$}. Only the patch size applied around anchor points varies: the higher, the greater the number of points that are activated during the fine pass. The first row represents the predictions, while the second row depicts the associated binary masks. All white points outside the mask have a zero prediction. The model only considers active points in the mask.
    }
    \label{fig:sparse_val_fine_ps}
\end{figure*}

\subsection{Anchor threshold}

We also vary the anchor threshold $\tau$ at which a point is considered an anchor point (\autoref{tab:ablations_thresh_fine}). It is noteworthy that the distribution of activation scores for the points exhibits a relatively long tail. Many points have an activation threshold between 0 and 0.1, and at 0.1, already 71\% of the points have been filtered. If the threshold is too high, there is a risk that certain regions will not be densified, leading to a drop in IoU. If we densify the entire map, we lose the memory value of sparse inference. This tradeoff is exhibited in \autoref{tab:ablations_thresh_fine}.

\begin{table}[h]
    \centering
    \setlength{\tabcolsep}{3pt}
    \begin{tabular}{@{}l *{6}{c} @{}}
    \toprule
    \textbf{Threshold} $\tau$ & 0.0 & 0.1 & 0.2 & 0.3 & 0.4 & 0.5 \\
    $N_\textit{fine}$ & 40,000 & 2,839 & 2,212 & 1,868 & 1,627 & 1,430 \\
    Memory (MB) & 1577 & 493 & 479 & 472 & 468 & 466 \\
    \midrule
    \textbf{IoU} vehicle ($\uparrow$) & 38.1 & 38.1 & 38.0 & 37.9 & 37.7 & 37.5 \\
    \bottomrule
    \end{tabular}
        \caption{\textbf{Sparse evaluation of our model on the nuScenes \cite{caesar2020nuscenes} validation set} at resolution $224 \times 480$ without visibility filtering, using different threshold to activate anchor points before the fine pass. The reported memory is the maximum memory allocated without the backbone.}
    \label{tab:ablations_thresh_fine}
\end{table}

\begin{figure*}
    \centering
    \includegraphics[width=\textwidth, trim=12cm 0 11cm 0, clip]{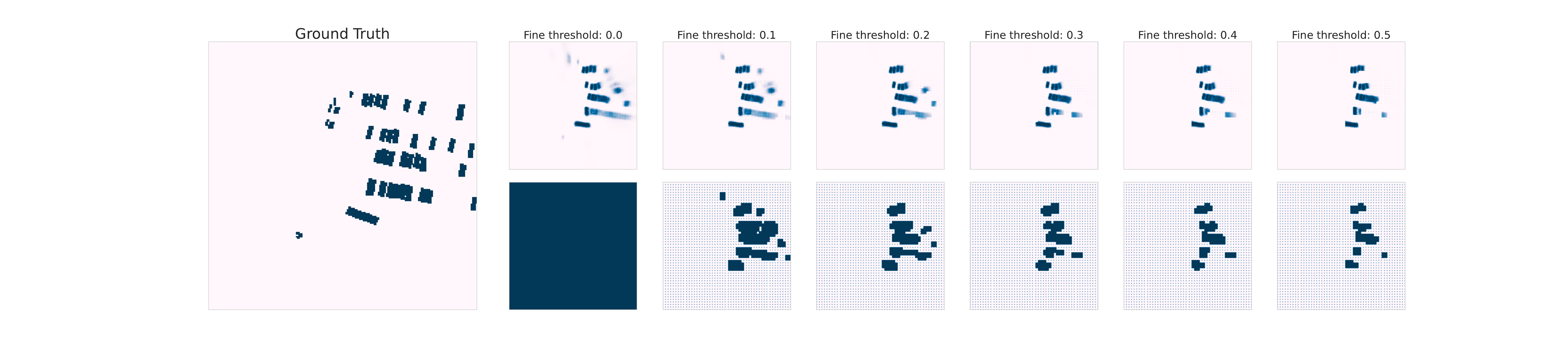}
    \caption{\textbf{Qualitative representation of different fine threshold $\tau_\textit{fine}$}. Only the threshold for activating anchor points varies: a lower threshold leads to a higher number of points designated as anchor points. The first row represents the predictions, while the second row depicts the associated binary masks. All white points outside the mask have a zero prediction. The model only considers active points in the mask.
    }
    \label{fig:sparse_val_fine_threshold}
\end{figure*}

Qualitatively, as the anchor point threshold increases, fewer points are considered active during the second pass, which may result in missing important regions in the BeV (see \autoref{fig:sparse_val_fine_threshold}). Conversely, a threshold set too low tends to regard too many points as significant, thereby losing the memory efficiency of the approach. Note that when the threshold is $\tau=0$, even if the image is sub-sampled, as the patch size is greater than half the spacing, we end up with an image that has made predictions over the entire BEV.
Considering the trade-off between the number of points considered, the anchor threshold and the final IoU, we have thus selected a default threshold factor of $\tau=0.1$, resulting in an IoU of 38.09 as indicated in \autoref{tab:sota:vehicle}.

\subsection{Sparse adaptative inference}

We also examine the influence of the reduction factor in the coarse pass by adjusting the kernel size to cover regions between two neighboring points (\autoref{tab:adaptative_fine}). This analysis is directly related to \autoref{fig:iou_vs_mem_vis2} in the paper. It demonstrates that subsampling 1/16 of the points is sufficient to achieve results similar to a model evaluating the entire grid in a single pass. Beyond a certain reduction factor, which is related to the size of the considered objects, performance decreases.

\begin{table}[t]
    \centering
    \resizebox{\linewidth}{!}{
    \begin{tabular}{@{}l @{\hspace{2mm}} c @{\hspace{2mm}} c @{\hspace{2mm}} c @{\hspace{2mm}}  c @{\hspace{2mm}}  c @{\hspace{2mm}}  c @{\hspace{2mm}} c @{}}
    \toprule
    \textbf{Subsample factor} $S_k$ & 1 & 2 & 4 & 8 & 16 & 32 & 64 \\
    \textbf{Densification size} $k_\textit{fine}$ & --- & 3 & 5 & 7 & 9 & 13 & 17 \\
    \midrule
    $N_\textit{coarse}$ & 40,000 & 19,880 & 10,000 & 4,900 & 2,500 & 1,255 & 625 \\
    $N_\textit{fine}$ & 0 & 1,270 & 1,640 & 1,917  & 2,017 & 2,245 & 2,349 \\
    Memory (MB) & 2379 & 1267 & 640 & 442 & 442 & 442 &  442 \\
    \midrule
     \textbf{IoU} vehicle ($\uparrow$) & 44.0 & 44.1 & 44.0 & 44.0 & 43.7 & 42.5 & 39.1 \\ 
    \bottomrule
    \end{tabular}
    }
        \caption{\textbf{Sparse evaluation of our model on the nuScenes \cite{caesar2020nuscenes} validation set} at resolution $224 \times 480$ without visibility filtering, using different reduction factors to sample regular coarse points on the BeV grid while adapting the kernel patch size. Backbone is an EfficientNet-b4 \cite{tan2020efficientnet} as in \autoref{fig:iou_vs_mem_vis2}.
        \label{tab:adaptative_fine}
        }
\end{table}

\begin{figure*}
    \centering
    \includegraphics[width=\textwidth, trim=12cm 0 11cm 0, clip]{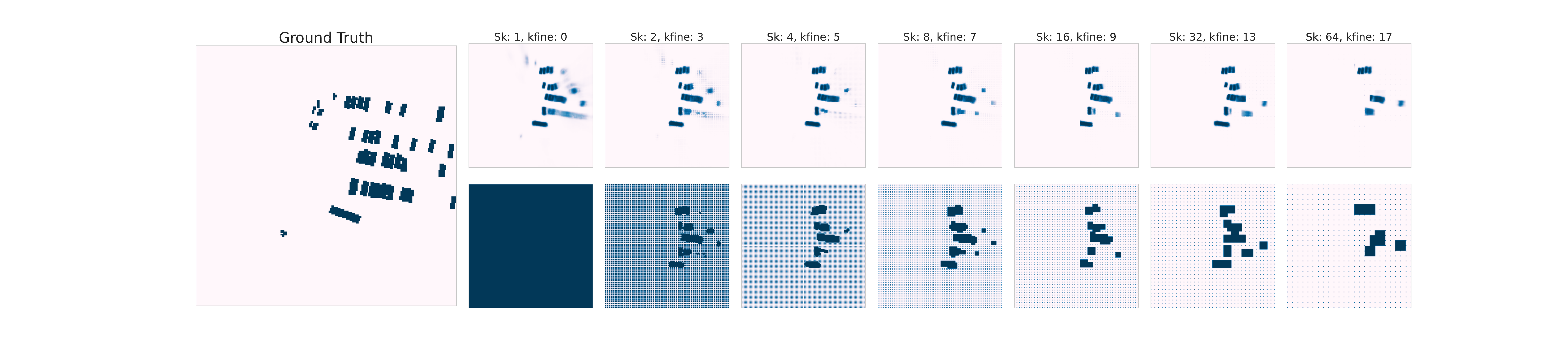}
    \caption{\textbf{Qualitative representation of different regular subsampling patterns} according to the reduction factor $S_k$ and with adaptation of the densification patch size $k_\textit{fine}$. The first row represents the predictions, while the second row depicts the associated binary masks. All white points outside the mask have a zero prediction. The model only considers active points in the mask.
    }
    \label{fig:sparse_val_adapt}
\end{figure*}

Qualitatively, it is observed that subsampling helps to correct predictions in certain areas of uncertainty (\autoref{fig:sparse_val_adapt}). However, if it is too high, there is a risk of missing important regions in the BeV. This is particularly the case when the factor is 64. Considering the trade-off between the number of points considered, and the final IoU, we have thus selected a default subsampling factor of $S_k=16$, resulting in an IoU of 43.73 as indicated in \autoref{tab:sota:vehicle}.

\subsection{LiDAR inference}

\begin{figure*}
    \centering
    \includegraphics[width=0.9\textwidth]{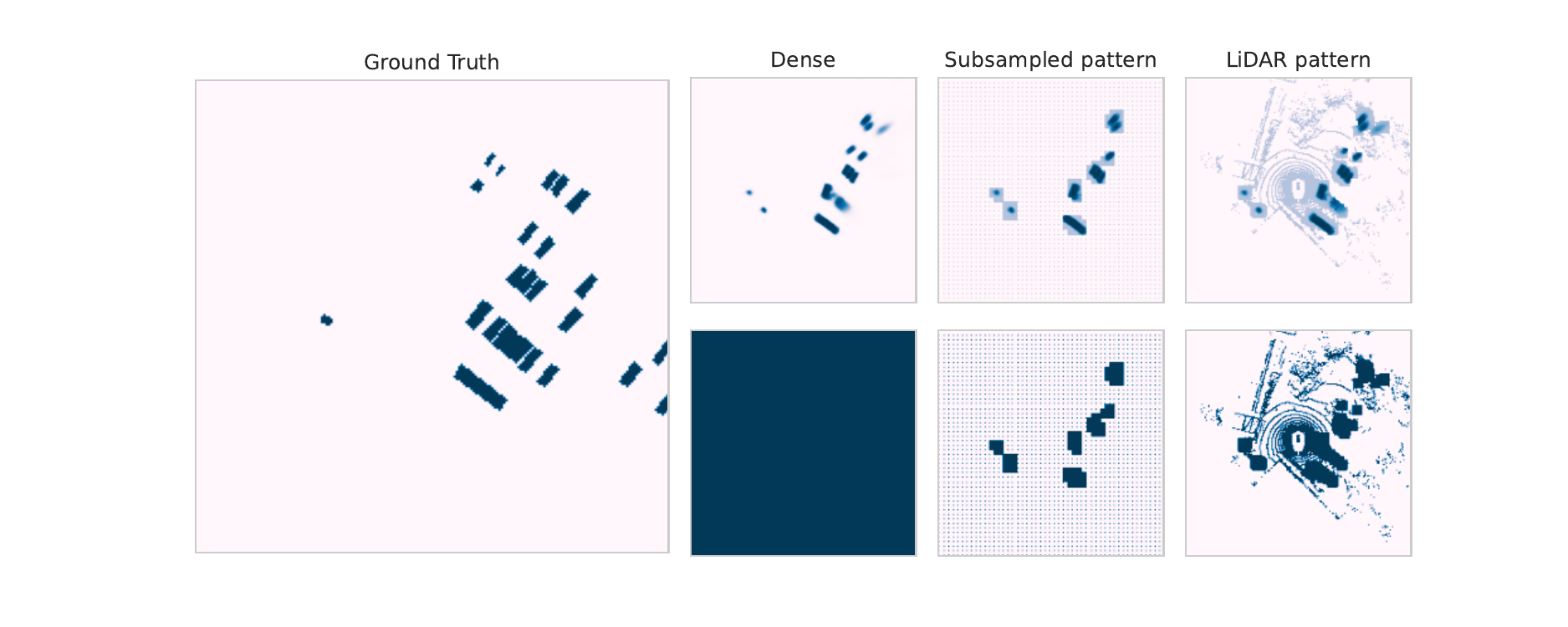}
    \caption{\textbf{Qualitative comparison of a subsampled pattern with a pattern initialized using LiDAR points.} The LiDAR points correspond to those from the sweep considered at the current timestep. The first row represents the predictions, while the second row depicts the associated binary masks. All white points outside the mask have a zero prediction. The model only considers active points in the mask.
    }
    \label{fig:comparison_lidar_pattern}
\end{figure*}

We discuss in the paper an initialization of the coarse pass using LiDAR points retrieved from a sweep (\autoref{fig:iou_vs_mem_vis2}). In the context of evaluation with a visibility filter, we demonstrate that this approach leads to better results than the standard approach or other sampling patterns (\autoref{tab:lidarvspatterns}). Visually, it can be observed that the LiDAR pattern activates more regions than the standard pattern, even though the same activation threshold for anchor points is used, see \autoref{fig:comparison_lidar_pattern}. This is mainly because LiDAR point locations are better candidates than regular locations, which have to rely on the receptive field of features to contain local information.

\begin{table}[h]
    \centering
    \resizebox{0.85\linewidth}{!}{
    \begin{tabular}{@{} l *{5}{@{\hspace{0.15cm}}c} @{}}
        \toprule
        \textbf{IoU ($\uparrow$) vehicle} & LiDAR & Random & Regular & Dense\\
        \midrule
        N$_\textit{point}$ (fine + coarse) & 5.4k & 2.7k & 2.7k & 40k \\
        PointBeV & 44.5 & 42.2 & 43.7 & 44.0 \\
         \bottomrule
    \end{tabular}
    }\caption{\textbf{Comparison of different sub-sampling patterns}, showing that the lidar pattern leads to the best results. Models are trained using EfficientNet-b4 with low visibility filtering.}
    \label{tab:lidarvspatterns}
\end{table}

\section{Qualitative examples}
\label{sec:supp:qualitative}

\begin{figure*}
    \centering
    \includegraphics[width=\textwidth]{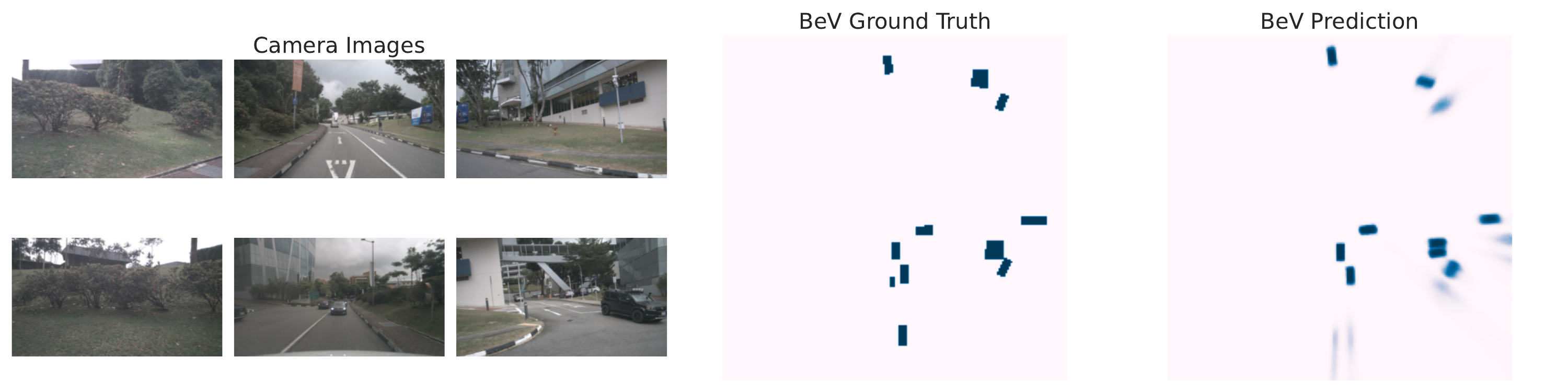}
    \caption{\textbf{Qualitative results of \acro{}'s prediction on a random sample from the nuScenes validation (not cherry-picked).} The model inputs are the six cameras displayed on the left, respectively the front-left camera, front camera, front-right camera, back-left camera, back camera, and back-right camera. The ground truth and then the prediction are displayed.}
    \label{fig:cameras_sun2}
\end{figure*}

\begin{figure*}[ht!]
    \centering
    \begin{subfigure}[b]{\textwidth}
        \includegraphics[width=\textwidth]{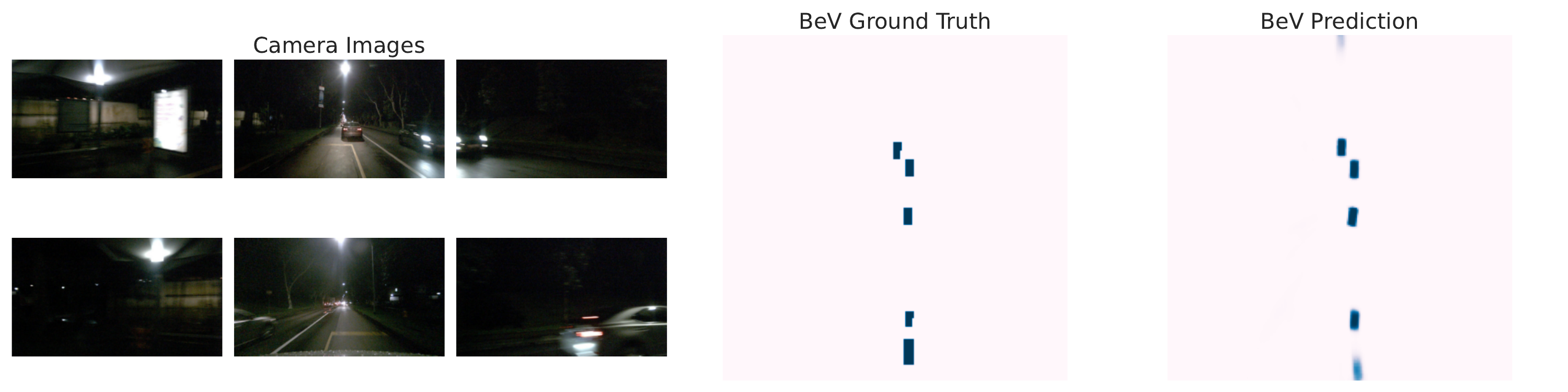}
        \label{fig:cameras_night}
    \end{subfigure}
    \hfill %
    \begin{subfigure}[b]{\textwidth}
        \includegraphics[width=\textwidth]{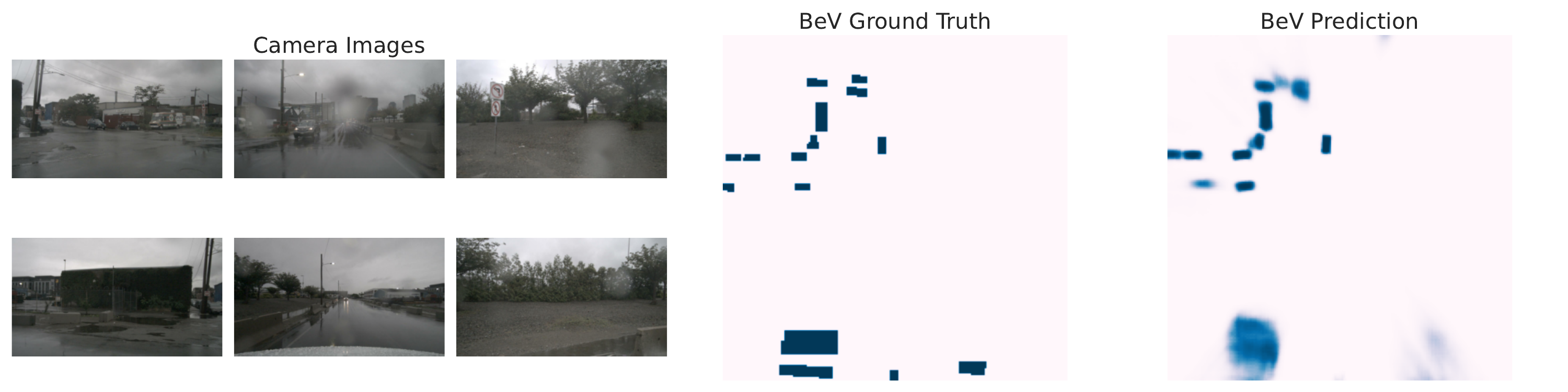}
        \label{fig:cameras_rain}
    \end{subfigure}
    \hfill %
    \begin{subfigure}[b]{\textwidth}
        \includegraphics[width=\textwidth]{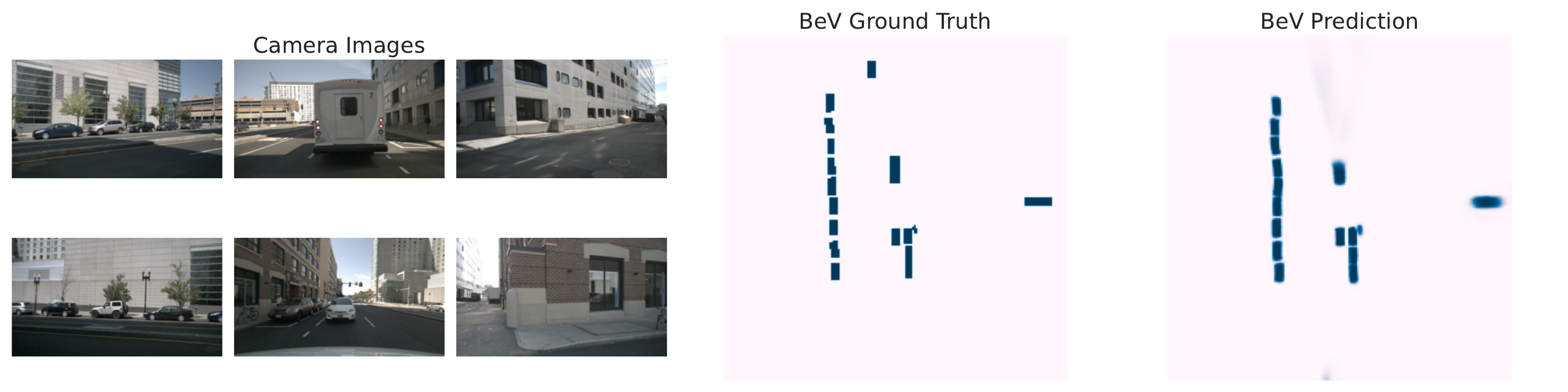}
        \label{fig:cameras_sun}
    \end{subfigure}
    \caption{\textbf{Qualitative results of \acro{}'s prediction on a nighttime, rainy and sunny sample from the nuScenes validation.} The model inputs are the six cameras displayed on the left, respectively the front-left camera, front camera, front-right camera, back-left camera, back camera, and back-right camera. The ground truth and then the prediction are displayed.}
    \label{fig:night_rain_sun}
\end{figure*}

We present visualizations of \acro{} vehicle occupancy map predictions on the nuScenes validation set (without visibility filtering) in (\autoref{fig:cameras_sun2}) and with various lighting and weather conditions (nighttime, rainy weather, and clear weather) in \autoref{fig:night_rain_sun}.

\end{document}